\documentclass{article}
\pdfoutput=1

\usepackage[dvipsnames,table]{xcolor}

\definecolor{qual-fig-green}{RGB}{0,144,11}
\definecolor{qual-fig-red}{RGB}{238,0,0}
\definecolor{qual-fig-purple}{RGB}{153,51,255}
\definecolor{cvprblue}{rgb}{0.21,0.49,0.74}
\definecolor{citecolor}{HTML}{0071BC}
\definecolor{linkcolor}{HTML}{ED1C24}

\usepackage[preprint,nonatbib]{neurips_2023}

\usepackage[utf8]{inputenc}
\usepackage[T1]{fontenc}
\usepackage{url}
\usepackage{booktabs}
\usepackage{amsfonts}
\usepackage{amsmath}
\usepackage{nicefrac}
\usepackage{microtype}
\usepackage{lipsum}
\usepackage{graphicx}
\usepackage{wrapfig}
\usepackage[size=small]{caption}
\usepackage{mathtools}
\usepackage{amssymb}
\usepackage{amsthm}
\usepackage[algoruled,boxed,lined,noend]{algorithm2e}
\usepackage{adjustbox}
\usepackage{float}
\usepackage{pifont}
\usepackage{subcaption}
\usepackage{tcolorbox}
\usepackage{pgfgantt}
\usepackage{multirow}
\usepackage[superscript,biblabel]{cite}
\usepackage[inline]{enumitem}
\usepackage{array}
\usepackage{tabularx}    
\usepackage{geometry}     
\usepackage{booktabs}  
\usepackage{hyperref}
\usepackage{hhline} 
\usepackage{multicol}

\usepackage[capitalize]{cleveref}
\usepackage{enumitem}
\usepackage{longtable}
\usepackage[nolists,nofigures,notables]{endfloat} 

\let\normallongtable=\longtable
\renewcommand{\longtable}{\clearpage\normallongtable}

\newcommand{\cmark}{\ding{51}}

\newcommand{\figref}[1]{Fig.~\ref{#1}}
\newcommand{\tabref}[1]{Tab.~\ref{#1}}

\newif\ifshowmarkup
\showmarkuptrue

\DeclareRobustCommand{\markup}[2][black]{%
  \ifshowmarkup
    {\begingroup\color{#1}#2\endgroup}%
  \else
    \ignorespaces 
  \fi
}

\DeclareRobustCommand{\nomarkup}[2][black]{%
  \ifshowmarkup
    \ignorespaces 
  \else
    {\begingroup\color{#1}#2\endgroup}%
  \fi
}

\newcolumntype{L}[1]{>{\raggedright\arraybackslash}p{#1}}

\title{Agentic Systems in Radiology: Design, Applications, Evaluation, and Challenges}
\author{\normalfont
Christian Bluethgen$^{1,7\ddagger}$
\quad Dave Van Veen$^{2}$
\quad Daniel Truhn$^{3,4}$ 
\quad Jakob Nikolas Kather$^{5}$ \\
\quad Michael Moor$^{6}$
\quad Ma{\l}gorzata Po{\l}acin$^{1}$
\quad Akshay Chaudhari$^{7,8,9}$
\quad Thomas Frauenfelder$^{1}$ \\
\quad Curtis P. Langlotz$^{7,8,9}$
\quad Michael Krauthammer$^{10}$
\quad Farhad Nooralahzadeh$^{10,11}$ \\
\vspace{3mm}
\\
$^1$Diagnostic and Interventional Radiology, University Hospital Zurich, University of Zurich, Zurich, Switzerland \\
$^{2}$HOPPR, Menlo Park, CA 94025, USA \\
$^{3}$Lab for AI in Medicine, University Hospital Aachen, Aachen, Germany \\
$^{4}$Diagnostic and Interventional Radiology, University Hospital Aachen, Aachen, Germany \\
$^{5}$Else Kroener Fresenius Center for Digital Health, Faculty of Medicine and University Hospital Carl Gustav Carus, \\ TUD Dresden University of Technology, 01307 Dresden, Germany \\
$^{6}$Department of Biosystems Science and Engineering, ETH Zurich, Basel, Switzerland \\
$^{7}$Center for Artificial Intelligence in Medicine and Imaging, Stanford University \\
$^{8}$Department of Radiology, Stanford University \\
$^{9}$Department of Biomedical Data Science, Stanford University \\
$^{10}$Department of Quantitative Biomedicine, University of Zurich, Zurich, Switzerland \\
$^{11}$ Institute of Computer Science, Zurich University of Applied Sciences, Zurich, Switzerland \\
}

\begin{document}

\maketitle
\thispagestyle{empty}
    \ifshowmarkup
    \else
    \fi
\begin{abstract}
Building agents, systems that perceive and act upon their environment with a degree of autonomy, has long been a focus of AI research. This pursuit has recently become vastly more practical with the emergence of large language models (LLMs) capable of using natural language to integrate information, follow instructions, and perform forms of "reasoning" and planning across a wide range of tasks. With its multimodal data streams and orchestrated workflows spanning multiple systems, radiology is uniquely suited to benefit from agents that can adapt to context and automate repetitive yet complex tasks. In radiology, LLMs and their multimodal variants have already demonstrated promising performance for individual tasks such as information extraction and report summarization. However, using LLMs in isolation underutilizes their potential to support complex, multi-step workflows where decisions depend on evolving context from multiple information sources. Equipping LLMs with external tools and feedback mechanisms enables them to drive systems that exhibit a spectrum of autonomy, ranging from semi-automated workflows to more adaptive agents capable of managing complex processes. This review examines the design of such LLM-driven \textit{agentic} systems, highlights key applications, discusses evaluation methods for planning and tool use, and outlines challenges such as error cascades, tool-use efficiency, and health IT integration.
\nomarkup{
\paragraph{Essentials}
\begin{itemize}
    \item LLM-driven agentic systems extend beyond simple text generation by using tools and feedback loops to automate multi-step radiology task.
    \item Such systems can function at different levels of agency, from structured workflows with limited decision-making to more adaptive agents with broader autonomy.
    \item Radiology departments can be conceptualized as dynamic environments that LLM-driven agentic systems observe, analyze, and act within.
    \item Evaluation must extend beyond single-task success to planning, reasoning, execution quality, reliability, efficiency, and safety.
    \item Key challenges include error propagation, human–AI coordination, regulatory compliance, and integration with existing health IT infrastructure.
\end{itemize}

\paragraph{Key Words} Large language models, Artificial Intelligence, Agent, Workflow, Automatization

\paragraph{Summary Statement} Large language model-driven systems ranging from controlled LLM-assisted workflows to more autonomous agents hold strong promise for orchestrating complex radiology tasks. Realizing this potential requires thoughtful integration, rigorous evaluation, and a clear understanding of associated challenges.
} 

\end{abstract}
\footnotetext{$^\ddagger$Corresponding author.}

\newpage
\setcounter{page}{1}
\pagestyle{plain}
\setcounter{section}{0}
\newpage

\begin{multicols}{2}
\section{Introduction}
\label{sec:intro}

Radiologists and their teams coordinate patients, operate scanners, interpret images, integrate clinical data, and communicate results. This multifaceted workflow demands adaptable problem-solving and frequent context switching; combined with rising imaging requests and a relative radiologist shortage, it contributes to cognitive overload and diagnostic delay~\cite{kwee2025workload,yahyavi2025ajr,rozenshtein2024ajr}.

AI is increasingly seen as a way to help manage this complexity~\cite{han2024randomised}, but many current implementations remain narrow-scoped and poorly integrated into clinical workflows. Large language models (LLMs) like GPT-5, and their multimodal variants, stand out for their ability to flexibly handle tasks specified in natural language~\cite{bhayana2024chatbots,van_veen_adapted_2024,Tu2025towardsconversational,McDuff2025towardsddx} at unprecedented accessibility. However, their effectiveness remains limited in real-world radiology, where tasks often involve multiple steps that unfold over time. When LLMs are used in isolation or called only once, as is typical in many current applications, they cannot adapt based on new information emerging during response generation.

LLM-driven \textit{agentic} systems address these limitations by embedding one or more LLMs within a framework in which LLMs can generate plans and select actions to iteratively interact with their environment~\cite{weng_llm_nodate,russel2020aimodernapproach}. In radiology, such systems could manage multi-step tasks that involve retrieving patient context, orchestrating specialized models, consulting external resources like guidelines, and synthesizing context-rich outputs like radiology reports. The field's data-rich, dynamic nature makes it well-suited for agentic approaches, but its complexity and clinical stakes require rigorous evaluation before deployment.

This review outlines the technical foundations of LLM-based agents, frames radiology as an agent environment and explores potential application, reviews methods for evaluating agent performance, highlights key challenges to clinical deployment, and considers future directions. Our objective is to illustrate to radiologists, researchers, and developers the potential of LLM-based workflows and agents to support complex, real-world radiological tasks.

\section{Technical Foundations of LLM-based Agents}
\label{sec:foundations}

\begin{figure*}
    \centering
    \includegraphics[width=1.0\linewidth]{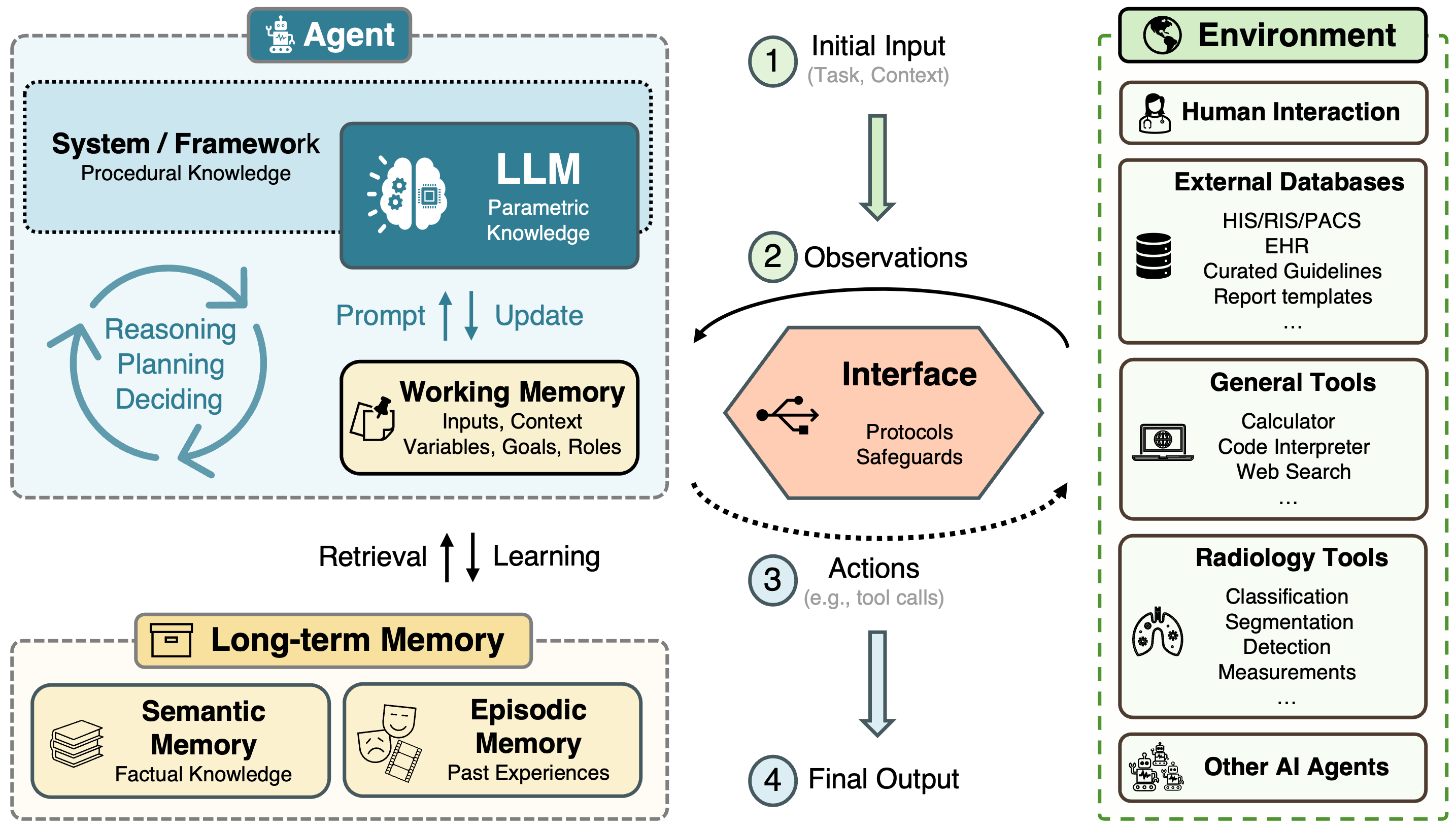}
    \caption{\textbf{Conceptual architecture of a radiology-focused LLM-based agent.} An initial input (1) provides the task and context. The agent then enters a cycle of obtaining observations (2), reasoning and planning over context, and performing actions (3) on the environment, such as tool calls or database queries. This cycle continues until a final output (4) is produced. The agent comprises an LLM, a framework, and a \textit{working memory}. An optional agent-owned \textit{long-term memory} stores \textit{episodic} (past interactions) and \textit{semantic} (factual knowledge) information to support retrieval and learning. The agent interacts with its \textit{environment} (green box), including external systems (e.g., HIS/RIS/PACS, EHR, databases), general and radiology-specific tools, humans, and other AI agents, via defined interfaces (e.g., Model Context Protocol (MCP), Agent-to-Agent (A2A)) and safeguards (e.g., PHI redaction, input validation). AI: Artificial Intelligence. EHR: Electronic Health Record. HIS: Hospital Information System. LLM: Large Language Model. PACS: Picture Archiving and Communication System. PHI: Protected Health Information. RIS: Radiology Information System.}

    \label{fig:architecture}
\end{figure*}

\begin{tcolorbox}[sharp corners,boxrule=0.4pt,colback=gray!8]{
    \textbf{What is an agent?} An \textit{agent} is an entity that perceives (through \textit{sensors}) and acts (through \textit{actuators}) on an \textit{environment}\cite{russel2020aimodernapproach}.
    LLM-based agents (\figref{fig:architecture}) run an LLM with access to external \textit{tools} in a \textit{loop} with some degree of autonomy in deciding \emph{which}, \emph{when}, and \emph{how} actions are executed to pursue a goal.
    }
\end{tcolorbox}

Here, we use "agentic" to refer to LLM-driven systems exhibiting goal-directed, feedback-adaptive behavior under limited supervision, including more autonomous \textit{agents} acting in open-ended settings, and less autonomous \textit{workflows} following predefined multi-step structures while making constrained, feedback-informed decisions within those boundaries. \markup{This distinction highlights differences in control, supervision, and adaptability, while recognizing that in practice, such systems often blend elements of both and that agency and autonomy lie on a spectrum\cite{effagents2024anthropic}.}

\subsection{LLMs as Agent Cores} 
\label{sec:agentcores}

Agents include a core component that transforms observations into actions, answering the question "Given the available information, what should the system do next?".
\markup{Earlier approaches to this problem relied on logical rules and structured representations (\textit{symbolic agents}), mapping observations directly to actions (\textit{reactive agents})~\cite{russel2020aimodernapproach,xi2025rise}, or reinforcement learning (\textit{RL agents}) that learn through interactions with (and reward signals from) their environment~\cite{sutton2018reinforcement,silver2021reward}. While RL has shown promise in narrow domains, its use in healthcare remains limited~\cite{yu2021reinforcement} as reward design is difficult, and trial-and-error learning can be unsafe or impractical.}

Unlike earlier agent designs\nomarkup{~that rely on manually defined rules or policies learned through trial-and-error feedback (as in reinforcement learning, RL)}, LLM-based agents use natural language prompts\markup{,}\nomarkup{~and}~intermediate feedback from their environment\markup{, and self-evaluation}~to guide decisions.
The strength of LLMs---and by extension, large multimodal models with LLM components---comes from pretraining on massive, diverse datasets~\cite{bluethgen2025best,paschali2025foundation}, which gives them broad general knowledge and the ability to process and produce language, follow instructions, reason and plan (to some extent)~\cite{ouyangTrainingLanguageModels2022,brown2020language,weiChainofThoughtPromptingElicits2023,hazra2025largelanguagemodelsreason,webb2024improvingplanninglargelanguage,kambhampati2024llmscantplanhelp}, and utilize tools and memory~\cite{schickToolformerLanguageModels2023,wang2024exploring,xi2025rise,nooralahzadeh2024explainablemultimodaldataexploration}.

This combination of abilities and general knowledge marks a qualitative shift:\markup{~for the first time, it has become practically feasible to build powerful AI agents for a wide range of domains. LLM-based agents are already being deployed across industries such as customer service, software development, and supply chain management, where they demonstrate adaptability and open-ended problem-solving beyond what earlier agent architectures could achieve~\cite{pwc_ai_agent_survey_2025}.}\nomarkup{~LLM-based agents have become practically feasible and are already deployed across industries for applications such as customer service and software development, where they demonstrate novel adaptability and open-ended problem-solving~\cite{pwc_ai_agent_survey_2025}.} These properties make them particularly interesting for complex, dynamic domains such as radiology.

\subsection{Environment, Tools and Actions}
\label{sec:environment}

An environment includes everything outside the agent that it can observe or influence through actions~\cite{russel2020aimodernapproach}. In radiology, this may include imaging devices, hospital and radiology IT systems, EHRs, and human stakeholders involved in clinical and administrative workflows.

To navigate an environment and produce useful decisions and outputs, an agent must connect its inner processes to external, "real-world" signals---a process known as \textit{grounding}~\cite{harnad1990grounding,bajaj2023grounding,liu2023grounding}. For example, describing a left lower lobe consolidation requires connecting the image content to the text sequences "left lower lobe" and "consolidation" (visual grounding~\cite{xiao2024visualgrounding}). These inputs (such as retrieved documents or images) may be handled directly by the (optionally multimodal) LLM or with help from external tools.

Tools are resources in the environment that agents use to \textit{sense} and \textit{act on} the environment. General tools include search engines, calculators, and code interpreters. In radiology, such tools could offer access to databases like PubMed or to specialized models (e.g., for segmentation). Tool effectiveness depends on alignment with the task and context, making tool optimization a key priority~\cite{effagents2024anthropic,huyen2025ai}. 

Tools fall into three broad categories: \markup{First,} tools for \textit{accessing dynamic or specialized knowledge}\markup{~to help agents move beyond static training data to retrieve up-to-date, patient-specific, or task-specific relevant information that would be impractical to include for every query (e.g., including lung cancer follow-up guideline texts during an abdominal MRI reporting workflow). Second}, tools that \textit{augment information processing}\markup{~to support tasks that remain difficult for LLMs, such as symbolic logic, math, or specialized vision tasks. Examples include segmentation models, dose calculators, or anatomical landmark detection to obtain measurements. Third,}
  \ifshowmarkup
  \else
    , and 
  \fi
tools that enable \textit{acting on the environment}\markup{~to allow agents to flag priority cases, schedule appointments, or communicate with remote monitoring devices}.

Protocols are emerging to standardize how agents use tools and interact with each other~\cite{yang_survey_2025}. One example is Anthropic's open Model Context Protocol (MCP), which defines a shared format for tool descriptions, requests, and responses. Instead of relying on custom application programming interfaces (APIs) that define explicit protocols how systems interact, agents can use MCP to flexibly discover and start using new tools while running. For instance, an agent can query an MCP-enabled EHR system without vendor-specific code, while still relying on established infrastructure. \markup{A2A (Agent-to-Agent) is another protocol handling secure, structured communication between agents themselves, allowing them to coordinate tasks, exchange data, and delegate subtasks in a standardized way.} Although promising, standards like MCP\markup{~and A2A} are still early and fragmented~\cite{yang_survey_2025}.

\markup{An agent's \textit{action space} is the set of external tools it can access and internal (LLM-native) actions it can perform (e.g., reasoning, summarization)~\cite{sumers2023cognitive}. Core agent functions include deciding \textit{when} and \textit{how} to act (e.g., providing the right input parameters at the right time, correctly parsing the returned output). 
Frameworks like LangChain~\cite{langchain2025}, DSPy~\cite{khattab2024dspy}, and HuggingFace's smolagents~\cite{smolagents2025} can help manage this logic. The action space can be expanded through LLM fine-tuning, loosening constraints, enhancing tools, or even enabling agents to create tools themselves~\cite{woelflein2025toolmaker}.}

\subsection{Goals, Reasoning and Planning}
\label{sec:goalsreasoningplanning}


LLMs pursue goals specified in natural language, which can leave room for ambiguity that can lead to unwanted outputs\markup{---behavior analogous to specification gaming \cite{amodei2016concrete,krakovna2020specification}}. For example, an instruction to "quickly complete radiology reports to improve turnaround time" may be interpreted as prioritizing speed over completeness, yielding terse or incomplete reports. Defining composite requirements for accuracy, completeness, and clinical appropriateness reduces this risk.

When faced with challenging requests, agents can apply \textit{reasoning}, which in the context of LLMs usually involves generating intermediate steps~\markup{(i.e., chain-of-thought reasoning)} to work through a problem systematically rather than jumping directly to conclusions\cite{weiChainofThoughtPromptingElicits2023}, and \textit{planning}, which constructs a sequence of actions expected to achieve the goal~\cite{russel2020aimodernapproach}. In radiology, this mirrors how a radiologist first organizes findings (reasoning) and then decides which prior studies and guidelines to consult, or which measurements to obtain, and in what order (planning). For example, interpreting a chest CT with multiple pulmonary nodules might involve (1) cataloging each nodule's characteristics, (2) comparing findings to prior imaging, (3) considering the patient's history and differential diagnoses, and (4) synthesizing recommendations. LLM-based agents can follow a similar structure by explicitly writing out their chain of thought before conclusions, typically yielding more accurate results than attempting a full assessment in one step\cite{weiChainofThoughtPromptingElicits2023}. 
\markup{In multimodal agents, this process may go beyond language: an agent might reason directly over images by generating predicted visual sequences~\cite{xu2025visual} or operate within a latent space before producing a final output~\cite{hao_training_2024}.}

A straightforward approach to handle complex requests is either break the task into subtasks manually or use a separate external planning systems to orchestrate the work~\cite{huyen2025ai}. Letting the model "think" longer while generating (\textit{test-time scaling}) can also help~\cite{wang2024exploring,geiping2025scalingtesttimecomputelatent}. \markup{Prompting strategies that expose or explore intermediate steps can raise accuracy further, for example by asking it to show its reasoning (\textit{chain-of-thought})~\cite{weiChainofThoughtPromptingElicits2023}, exploring several solution paths and keeping the majority answer (\textit{self-consistency})~\cite{wang2023selfconsistency}, or searching over branching ideas (\textit{tree-of-thoughts})~\cite{yao2023tree,koh2024tree}. Recent large reasoning models (e.g., OpenAI's o3 or DeepSeek-R1) are specifically trained to reason (although traces are not always returned to the user), achieving high performance on complex (non-medical) tasks~\cite{openai_o3_2025,Guo2025DeepSeek}.

Beyond these LLM-centered techniques, LLMs embedded in agentic systems operate in loops, mixing reasoning with actions and feedback (reason-act-observe loops, \textit{ReAct})~\cite{yao2023react} or separate planning from evidence gathering (\textit{ReWOO})~\cite{xu2023rewoo}; some add self-critique to improve the next attempt (\textit{Reflexion})~\cite{shinn2023reflexion}.}

\subsection{Context, Memory and Learning}
\label{sec:context}

Context is the information available to the LLM when processing a request, including user instructions, conversation history, and any external information such as tool outputs~\cite{schmid2024context}. Just as holistic image interpretation needs the right clinical information at the right time, agents require careful \textit{context engineering}---providing the LLM with optimal information in the most effective format and timing\cite{schmid2024context,mei2025contextengineering}. \markup{ Since LLMs have limited context windows (i.e., the information they can process at once), techniques like summarization help optimize what gets included\cite{li2024promptcompression}.} 

Beyond optimizing existing context, systems can enrich it dynamically through retrieval-augmented generation (RAG), which queries a (trusted) knowledge source (e.g., a database) and provides the LLM with results. RAG may augment each call automatically, be run as a tool, or operate agentically with dynamic retrieval and processing. In radiology, this can mean fetching prior reports, templates, or guidelines, analogous to how radiologists retrieve task-relevant additional information.\markup{~The value of RAG depends on both the quality of the source information and the performance of the retrieval system\cite{zakkaAlmanacRetrievalAugmentedLanguage2024,zheng2025miriad,arasteh2024radiorag}.}

LLMs by themselves are stateless, meaning each new response depends only on the current input. Additional memory systems maintain continuity across interactions\cite{huyen2025ai,sumers2023cognitive}. The LLM's \textit{internal knowledge} is fixed at training and not reliable for up-to-date, factual information\cite{truhn2023large}.\markup{~Together with prompts or configurations provided by the serving framework, it forms the agent's "procedural memory" \cite{sumers2023cognitive}. Updating this knowledge requires model fine-tuning or framework changes.}
\textit{Short term memory} functions as the agent's working space. It holds the current conversation state, including recent observations from the environment and outputs from tools.\markup{~This may be limited to the LLM's context window or managed via external structures across multiple calls. Known as \textit{state management}, this process involves tracking, updating, and discarding elements like conversation history or user preferences to avoid context drift or degradation (i.e., the gradual loss or distortion of relevant context that can compromise decision-making quality)\cite{hong2025contextrot}.}
~\textit{Long term memory} \markup{includes both general facts (semantic memory) and records of past actions (episodic memory). It }can store useful information for future reference, such as user preferences, successful response patterns, examples for learning, or guidelines (making it a frequent source for RAG systems~\cite{sumers2023cognitive}), but raises critical considerations of data governance and privacy when storing patient information for future use.

Some agent systems can improve over time by keeping track of past successes~\cite{liAgentHospitalSimulacrum2024a,schmidgallAgentClinicMultimodalAgent2024}\markup{(experiential learning).~Even without human feedback, \textit{self-evolving agents} can adapt their own skills, memory, and tools through rewards, imitation, or search across strategies~\cite{gao2025selfevolve}. Examples include agents that refine themselves by comparing against earlier versions ("self-play")~\cite{chen2024spin} or by generating tasks with built-in verification and learning from the outcomes~\cite{zhou2025self}}. For radiology, such improvement capabilities could enable agents to adapt to user preferences like preferred terminology, structural conventions of individual radiologists or departments -- a crucial feature for clinical adoption and integration into existing workflows~\cite{kimSeeingUnseenAdvancing2024}.

\subsection{Design Patterns for Agentic Systems}
\label{sec:design_patterns}

\begin{figure*}
    \centering
    \includegraphics[width=1\linewidth]{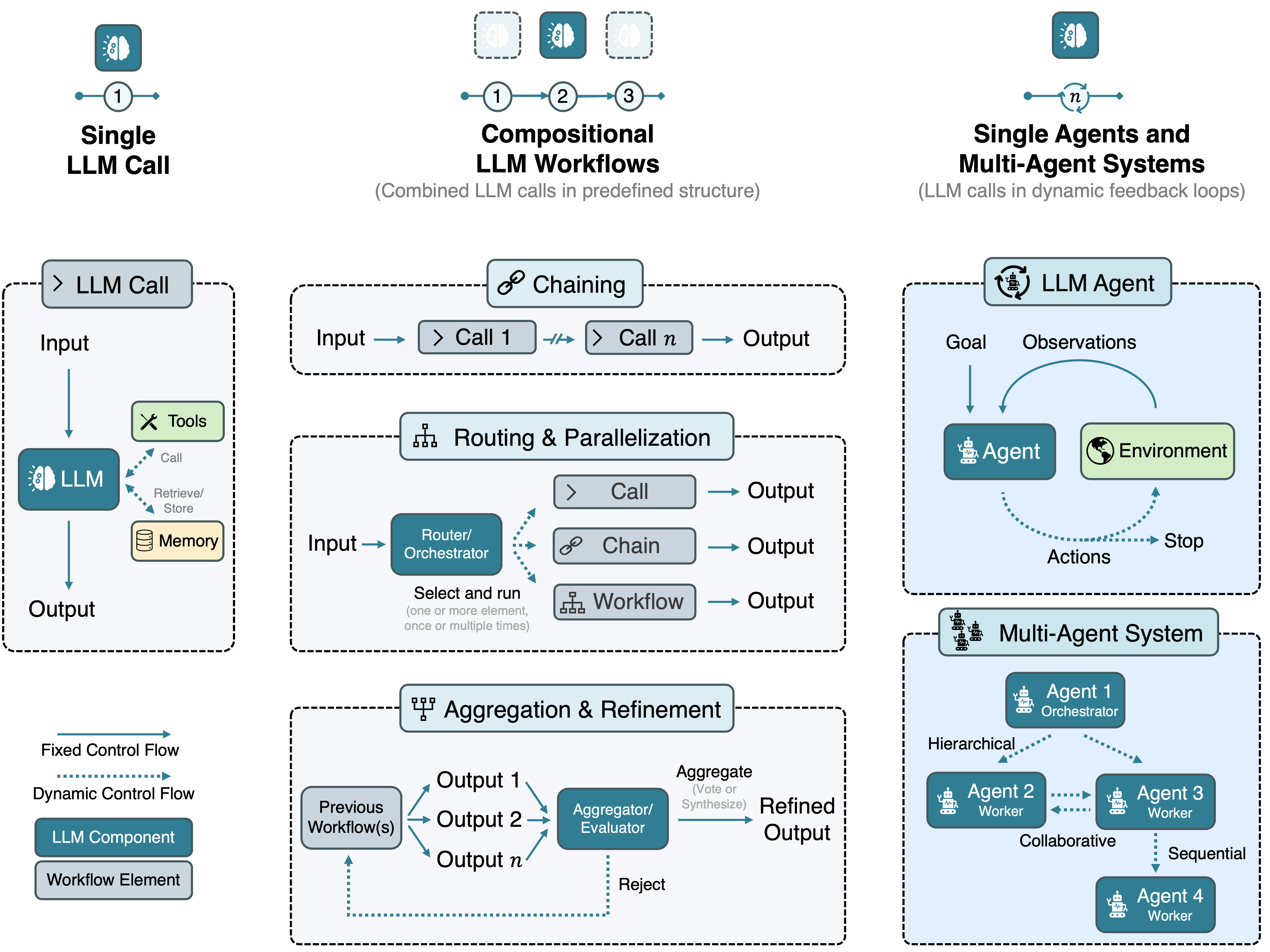}
    \caption{\textbf{Overview of building blocks and design patterns for LLM-based agentic systems}. The illustrated components are modular rather than mutually exclusive and can be combined to arbitrary complexity. \textbf{(Left column)} A single LLM call forms the basic building block, optionally reading from or writing to external tools or memory. \textbf{(Center column)} Multiple LLM calls can form workflows through (i) \textit{chaining} in fixed sequences, (ii) \textit{routing and/or parallelization} by an orchestrator LLM, or (iii) \textit{aggregation and refinement} by an evaluator that synthesizes or rejects results. \textbf{(Right column)}
    Agent systems extend this pattern: a single agent interacts with its environment through observation–reason–action loops, while multi-agent systems organize agents hierarchically (e.g., manager–subagent), collaboratively (specialized roles), or sequentially.
    LLM: large language model.}
    \label{fig:building_blocks}
\end{figure*}

LLM-driven systems can be organized by increasing level of complexity and autonomy: single LLM interactions, structured workflows, and systems with one or more autonomous agents. Each approach differs in how it handles \textit{control flow} -- the sequence and logic of executing actions required for a multi-step task\cite{huyen2025ai}.

\markup{Control flows can take several forms: sequential (steps executed in order), branching (different paths based on conditions), parallel (multiple simultaneous tasks), or looping (repeating actions until a condition is met).}
In conventional software, programmers predefine these flows. LLM-driven agentic systems, however, can generate them dynamically\cite{huyen2025ai}. 

Several useful design patterns have emerged (\tabref{tbl:workflow-agentic}, \figref{fig:building_blocks}). The fundamental building block is a single interaction with (or "call" to) an \textit{augmented LLM}, which may include tool or memory use (e.g., summarizing a provided report).

When a single call is insufficient, multiple calls can be combined into structured \textit{workflows} for multi-step tasks (e.g., retrieving prior data, fetching a template, then drafting a report). Common control flow patterns in such workflows include \textit{chaining} (using one output as the next input), \textit{routing} (selecting among several paths), and \textit{parallelization} (executing independent subtasks or repeated attempts with aggregation)\cite{effagents2024anthropic,wu2022ai}.

When control flows cannot be predefined easily (for instance, conducting a systematic literature research), a more autonomous \textit{agent} iterates (e.g., in reason-act-observe loops~\cite{yao2023react}) until a stop condition is met (goal reached, critical error, or budget exhausted). This enables more adaptive problem-solving.

\textit{Multi-agent systems} (MAS) coordinate multiple agents that delegate tasks, communicate, and share tools and memory. \markup{They can be organized in different ways (topologies), for example as peer-to-peer networks (all agents collaborate) or hierarchically (supervisor agents hand off tasks to specialized sub-agents)~\cite{wu2024autogen,zhang2025agentorchestra}. Another key design choice is between general-purpose agents with broad tool access and specialized agents focused on narrow tasks. While generalists simplify coordination, specialists often achieve higher tool-use accuracy and speed, especially under hierarchical delegation~\cite{effagents2024anthropic,zhang2025agentorchestra,wu2024autogen}.} MAS can outperform single-agent setups~\cite{zhang2025agentorchestra,chen2024autoagents} and exhibit complex group behaviors~\cite{jimenez2025swarm}, but greater autonomy also amplifies oversight and error-propagation concerns.

Choosing how to structure the system depends on the task and the desired balance between control and flexibility. LLM-based \textit{workflows} suit predictable, auditable tasks and are easier to maintain, with LLM decision-making autonomy limited to predefined points. Autonomous \textit{agents} offer greater adaptability for complex problems but require more error handling and oversight, especially in MAS. Practitioners recommend adding complexity only when necessary\cite{effagents2024anthropic}. This means starting with an augmented LLM call, using workflows when steps are known and control is critical (applying to many radiology tasks), and adopting agents when their flexibility justifies the extra cost and risk. Guardrails and evaluation depth should scale with system autonomy.

\definecolor{lightgray}{RGB}{240,240,240}

\begin{table*}[t]
    \centering
    \scriptsize
    \renewcommand{\arraystretch}{1.15}
    \setlength{\tabcolsep}{4pt}
    \begin{tabularx}{\textwidth}{p{3cm} X X}
    \toprule
    \textbf{Pattern} & \textbf{Core idea} & \textbf{When to use} \\
    \midrule

    \rowcolor{lightgray} \multicolumn{3}{l}{\textbf{Basic Building Block}} \\
    Augmented LLM call
    & Single LLM call enhanced with tools and memory.
    & Baseline choice; start here before adding complexity. \\

    \rowcolor{lightgray} \multicolumn{3}{l}{\textbf{Compositional Workflows}} \\
    Chaining
    & Break a task into a fixed sequence of LLM calls; each call is informed by previous output.
    & Tasks with a natural linear decomposition (e.g., outline $\rightarrow$ draft $\rightarrow$ full report). \\

    Routing
    & A router classifies the input, then dispatches to specialized workflows or models.
    & Tasks requiring one of several specialized models (e.g., call a workflow for CT analysis). \\

    Sectioning
    & Split input into independent complementary subtasks, solve in parallel, then aggregate.
    & Tasks requiring a foreseeable number of predefined steps (e.g., retrieving patient appointments from EHR and previous studies from RIS to create a summary). \\

    Voting
    & Run the same prompt multiple times and score/majority-vote results (self-consistency).
    & Safety checks, hallucination reduction. \\

    \rowcolor{lightgray} \multicolumn{3}{l}{\textbf{Agent Systems — Single-Agent}} \\
    ReAct Agent
    & An agent iterates in a \textbf{Re}ason-\textbf{Act}-Observe loop until a stop criterion is met.
    & Ill-defined or dynamic tasks where steps cannot be predetermined. \\

    Plan-and-Execute
    & An agent produces a multi-step plan and follows it step-by-step (usually single-agent).
    & Tasks that allow or require long-term planning. \\

    \rowcolor{lightgray} \multicolumn{3}{l}{\textbf{Agent Systems — Multi-Agent (MAS)}} \\
    Orchestrator–Workers
    & A central agent plans subtasks and delegates to worker agents, then merges results.
    & Broad, open-ended tasks, e.g., systematic research (e.g., Biomni\cite{huang2025biomni}) \\

    Evaluator–Optimizer
    & One agent drafts, another critiques and suggests fixes; loop until quality matches threshold.
    & Iteratively improving outputs, e.g. writer and reviewer agents taking turns to improve impression sections (RadCouncil)\cite{zeng2024radcouncil}. \\

    Peer-to-Peer (Swarm)
    & Multiple autonomous agents collaborate as equals without a central coordinator; involves communication (e.g., via shared memory).
    & Large-scale exploration, distributed ownership, or when central orchestration is a bottleneck\cite{jimenez2025swarm}.\\

    \bottomrule
    \end{tabularx}
    \caption{\textbf{Design patterns of LLM-based workflows and agentic systems.} The basic building block is an LLM with access to tools and memory. Workflows combine multiple LLM calls in sequence, with branching or parallel logic, optionally aggregating the outputs (e.g., by scoring). Agents add autonomy by iteratively reasoning, acting, and observing feedback from the environment in loops until a stop criterion is met (e.g., goal reached, iteration budget exhausted). MAS extend this by coordinating multiple agents that communicate, allocate or negotiate subtasks, and act sequentially or concurrently. LLM: Large language model. MAS: Multi-agent system.}
    \label{tbl:workflow-agentic}
\end{table*}

\section{Radiology as Environment for Agents}
\label{sec:rad_environment}

Viewing radiology as an "environment" clarifies what an agent can observe and act upon, highlighting challenges unique to radiology. The field's environment is complex and multimodal: it includes images, structured and unstructured EHR information, metadata from radiology and hospital information systems, and the speech, gestures, or written communication of clinicians and patients. These characteristics directly influence the design of agentic systems\cite{russel2020aimodernapproach}.

\markup{From a technical perspective, the radiology environment involves partial observability (e.g., incomplete patient data), a mix of episodic and longitudinal observations (single studies vs. follow-ups), and occasional real-time responsiveness (e.g., image-guided procedures). It is inherently multi-agent, encompassing radiologists, technologists, referrers, patients, and IT systems, with agents confronting missing data, evolving procedures, structured reporting, and cross-role coordination.}

\subsection{Radiology's Toolbox}
\label{sec:rad_toolbox}
Many tasks in radiology exceed what LLMs can handle in isolation. Agentic AI may hand off such tasks to specialized software (e.g., CAD tools), and other AI models like TotalSegmentator\cite{wasserthal2023totalsegmentator} for segmenting anatomical structures or foundation models\cite{paschali2025foundation} adapted for chest X-ray\cite{chenCheXagentFoundationModel2024,deperrois2025radvlm} or CT\cite{hamamci_developing_2025,blankemeier2024merlin} analysis and reporting.

Agentic tool use can significantly enhance task performance: For example, Ferber et al. demonstrate a boost from 30\% to 87\% accuracy over isolated LLM use~\cite{ferber2025development}. To achieve this, agents must "understand" what each tool does, when (and in which order) to use it, and how to interpret the output in the radiology context~\cite{zheng_can_2024}. For example, lung nodule assessment might involve calling CAD software and interpreting measurements returned in a structured format \markup{(e.g., JSON)}. An "agent-friendly" interface (e.g., MCP) helps by allowing the LLM to understand and communicate with the tools. For instance, RadFabric is an MCP-based multi-agent setup with specialized agents handling CXR analysis and report generation~\cite{chen2025radfabric}.

\subsection{Radiology-specific Knowledge Sources} 
\label{subsec:rad_knowledge_sources}

Radiologists often consult literature or reference cases to refine diagnoses. Similarly, LLM-based agents can query databases such as PubMed, Radiopaedia, the RSNA Case Collection to incorporate up-to-date information.

Standardized image reporting systems (BI-RADS, PI-RADS, LI-RADS, Lung-RADS) reduce variability, ground recommendations in evidence, and facilitate interdisciplinary communication~\cite{an_bi-rads_2019}, while broader frameworks such as TNM staging~\cite{uicc2025tnm9} and RECIST~\cite{eisenhauer2009recist1_1} integrate imaging with clinical context.
\markup{While some argue that LLMs could make natural language a universal interface in healthcare, reducing the reliance on fixed schemas~\cite{kather2024llminterface}, }\ifshowmarkup
    validated
 \else
    Validated
  \fi
ontologies offer human-readable and machine-computable representations of radiology concepts and relations, enabling semantic interoperability~\cite{filice_integrating_2019,chepelev_ontologies_2023}. For example, SNOMED CT encodes clinical concepts and their relations to support structured documentation and data exchange. RadLex focuses on imaging-specific concepts and supports standardization efforts such as the RSNA-ACR Common Data Elements and the RSNA-LOINC Radiology Playbook~\cite{vreeman_loinc_2018}. The Radiology Gamuts Ontology (RGO) formalizes radiological differential diagnoses~\cite{budovec2014informatics}. RadGraph-XL structures radiology data as \textit{knowledge graph}, where nodes represent clinical entities (e.g., "right upper lobe") and edges encode relations (e.g., "suggestive of")~\cite{delbrouck2024radgraph}.

Grounding the agent in structured medical knowledge guides it to operate within consistent, interpretable categories and reduces the risk of clinically ambiguous outputs or confabulations\cite{zakkaAlmanacRetrievalAugmentedLanguage2024,chang2024snomed}. For example, RadioRAG raises diagnostic performance on expert-curated tasks by up to 54\%~\cite{arasteh2024radiorag}.

\subsection{Radiology's Ecosystem}
\label{subsec:rad_infrastructure}

Radiology operates within a digital ecosystem that agents must interface with to access, process, and act on clinical data. This ecosystem includes the Hospital Information System (HIS) for patient administration and clinical history, the Radiology Information System (RIS) for radiology-specific tasks like scheduling and reporting, and the Picture Archiving and Communication System (PACS) for image storage and distribution. PACS is often complemented by a vendor-neutral archive (VNA) for long-term retention. As data volumes grow, data lakes and warehouses are increasingly adopted to support analytics and AI workflows~\cite{arnold2024fhir}.

Interoperability across these systems relies on standardized communication protocols. DICOMweb extends the DICOM standard, which governs the storage and transmission of images, structured reports, and segmentation data, into modern \markup{~RESTful web} APIs that allow communication using simple, standardized web requests, enabling scalable, network-based integration with agents. Fast Healthcare Interoperability Resources (FHIR) builds on the Health Level 7 (HL7) standard for clinical data exchange by defining modular resources\markup{~(e.g., "ImagingStudy") and by supporting semantic interoperability through established vocabularies like SNOMED CT, RadLex, and LOINC}~\cite{arnold2024fhir}.\markup{SMART on FHIR introduces secure OAuth2-based access control, while FHIR Subscriptions and FHIRcast provide real-time event updates, allowing an agent to be notified when a new study arrives or when a radiologist opens a case.}

\begin{figure*}
    \centering
    \includegraphics[width=1\linewidth]{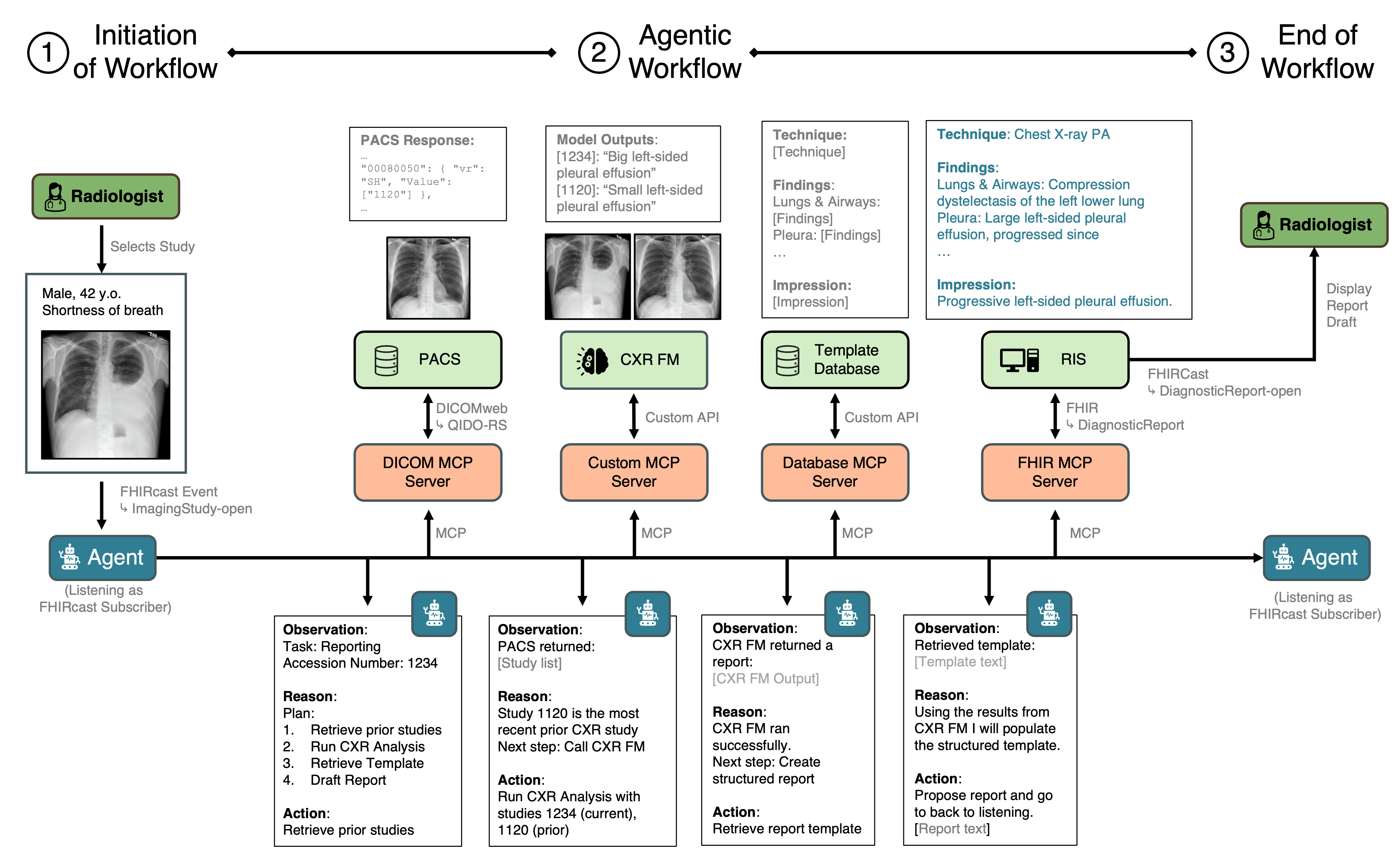}
    \caption{\textbf{Example of an agentic workflow for report drafting within traditional radiology infrastructure.} The workflow begins with a radiologist selecting a study (Step 1), triggering an FHIRcast event that notifies the agent. The agent plans the reporting task by retrieving prior studies from PACS through a DICOM MCP server (Step 2). It then calls a chest X-ray foundation model (CXR FM) via a custom MCP server to analyze the current and prior images. Once the model returns findings, the agent retrieves a structured report template from a template database via a database MCP server and populates it with model output. The structured draft report is sent to the radiology information system (RIS) using the FHIR protocol via an FHIR MCP server. The workflow concludes when the radiologist receives and reviews the draft report (Step 3). The agent continuously observes, reasons, and takes actions throughout the process via Model Context Protocol (MCP) interactions. DICOM: Digital Imaging and Communications in Medicine. CXR: Chest X-Ray. API: Application Programming Interface. FHIR: Fast Healthcare Interoperability Resources. FM: Foundation Model. MCP: Model Context Protocol. PACS: Picture Archiving and Communication System. RIS: Radiology Information System.}
    \label{fig:workflow_example}
\end{figure*}

Building on top of these protocols (optionally after wrapping them in MCP~\cite{ChristianHinge2025dicommcp}), developers can create complex agentic radiology applications while leveraging existing infrastructure for security and stability(\figref{fig:workflow_example}).

\section{Applications in Radiology}
\label{sec:applications}

While agentic AI has not entered radiology practice, early studies in other medical fields suggest potential for improved triage, decision-making, and efficiency\cite{collaco2025agenticAI}. Here, we walk through exemplary agentic solutions for radiology at varying degrees of autonomy.

\markup{
\subsection{Chest X-ray Consistency Checker}\label{example:cxrchecker}
When a chest radiograph is opened in the viewer, an agent monitors the dictation stream and concurrently retrieves prior chest films for comparison. As the radiologist begins their impression, the agent asynchronously checks for line- and tube placement, and comparison statements. If, for example, the dictation lacks a reference to a newly placed device detected in the image, the agent highlights this as a possible omission. A prompt appears: "Previous film (2 days ago) had no left CVC. Include new line in report?" Upon confirmation, the agent suggests templated language. All edits are logged (for posterity, personalization and future improvements), and the radiologist maintains control over final wording.
}


\subsection{Agentic Lung Cancer Screening Reporting}\label{example:lungcancer}

Lung cancer screening (LCS) programs are increasingly implemented, leading to greater workloads~\cite{strayer2023lcs}. Their structured, multi-step nature makes them ideal for workflow-based automation with limited autonomy. In this setting, an agent manages routine steps while allowing the radiologist to focus on interpretation.

When an exam is opened, the agent identifies the LCS scenario, retrieves validated workflow-specific instructions, and pulls patient data from PACS, RIS, and HIS, including prior CTs and smoking history. AI-driven image registration\cite{meng2025totalregistrator} aligns current and prior scans to support longitudinal nodule tracking in side-by-side comparison, a process that is typically time-consuming. A CAD model detects nodules, producing structured descriptors (size, texture, lobe location). After radiologist review, the agent assigns Lung-RADS categories, loads the appropriate report template, and drafts the report. Internal consistency checks are applied, and any radiologist edits are logged to personalize future drafts. Once signed, follow-up recommendations are communicated to the referring provider.

\markup{This workflow uses DICOMweb for image management, FHIR for structured clinical data exchange with the EHR, and RAG to pull guideline content from a curated database.}

\subsection{Agent‑Assisted Tutoring through Interactive Reporting}\label{example:education}

Interactive learning assistance in routine reporting illustrates a more autonomous, conversational workflow. In this setting, the agent engages flexibly with users rather than following a fixed structure. A resident drafting a report can consult an integrated tutor agent to clarify findings or better understand concepts. The agent answers free-text queries with targeted teaching, using curated cases and generative tools to create illustrative examples.

For example, if the resident asks about a pneumonic infiltrate near the right heart border on a chest radiograph, the agent explains the silhouette sign and generates paired synthetic images showing middle and right lower lobe \cite{bluethgen_visionlanguage_2024}. Once the resident is finished drafting, the agent checks for internal inconsistencies such as laterality mismatches and flags missing responses to relevant clinical questions. 

In this example, the agent can provide timely, specific feedback and flexibly adjust to the resident by retrieving additional content (e.g., through RAG) or generate illustrative examples on-the-fly (through a tool call). All interactions can be logged for longitudinal skill tracking.

\subsection{Enhanced Multidisciplinary Team Discussions}\label{example:mdt}

Multidisciplinary team discussions (MDT) rely on thorough preparation and documentation, as well as rapid access to additional information during the meeting. Often, radiologists manage the displayed content, manually retrieving additional imaging and pertinent records during the discussion. This scenario benefits from mix of predefined workflows and adaptive agentic assistance.

Before the meeting, an agent can assemble the case list by extracting information from communications, scheduling systems, and prior meeting notes, while verifying the completeness of imaging, pathology, and laboratory data. It then condenses relevant history, imaging findings, and treatment timelines into concise case summaries that are shared with participants in advance.

During the meeting, a domain-adapted automatic speech recognition (ASR) engine transcribes the discussion with high accuracy\cite{dabass2025asrllm}. 
The agent monitors the transcript to detect information requests, then queries the EHR to retrieve and display relevant data such as prior imaging, pulmonary function tests, or medications, all in real time.

After the meeting, the agent consolidates the transcript, the retrieved data, and the decisions recorded during the session. It generates a structured summary containing patient-specific conclusions, assigned responsibilities, and relevant supporting images or reports. This summary is stored in the EHR, shared with participants, and used to trigger follow-up actions such as scheduling, ordering tests, or notifying referring clinicians.

\subsection{Agent‑Driven Follow-up Scheduling}\label{example:scheduling}

Scheduling follow-up imaging is a repetitive but essential process that often involves multiple parties beyond radiology. An agent can receive a referral, determine the recommended interval based on guidelines and the prior exam date, and query hospital scheduling systems for equipment and staff availability. It can then contact the patient via phone or secure portal to propose available time slots and explain the clinical importance of the follow-up.

Once the appointment is accepted, the agent suggests an appropriate imaging protocol, seeks radiologist approval where needed, and finalizes the booking across RIS, PACS, and HIS. Notifications are sent to all stakeholders, and reminders are issued automatically. A rule-based check ensures correct protocolling for each referral type, and audit logs maintain a record of all actions for compliance.

\section{Evaluation of Radiology Agents}
\label{sec:evaluation}

\begin{figure*}
    \centering
    \includegraphics[width=1\linewidth]{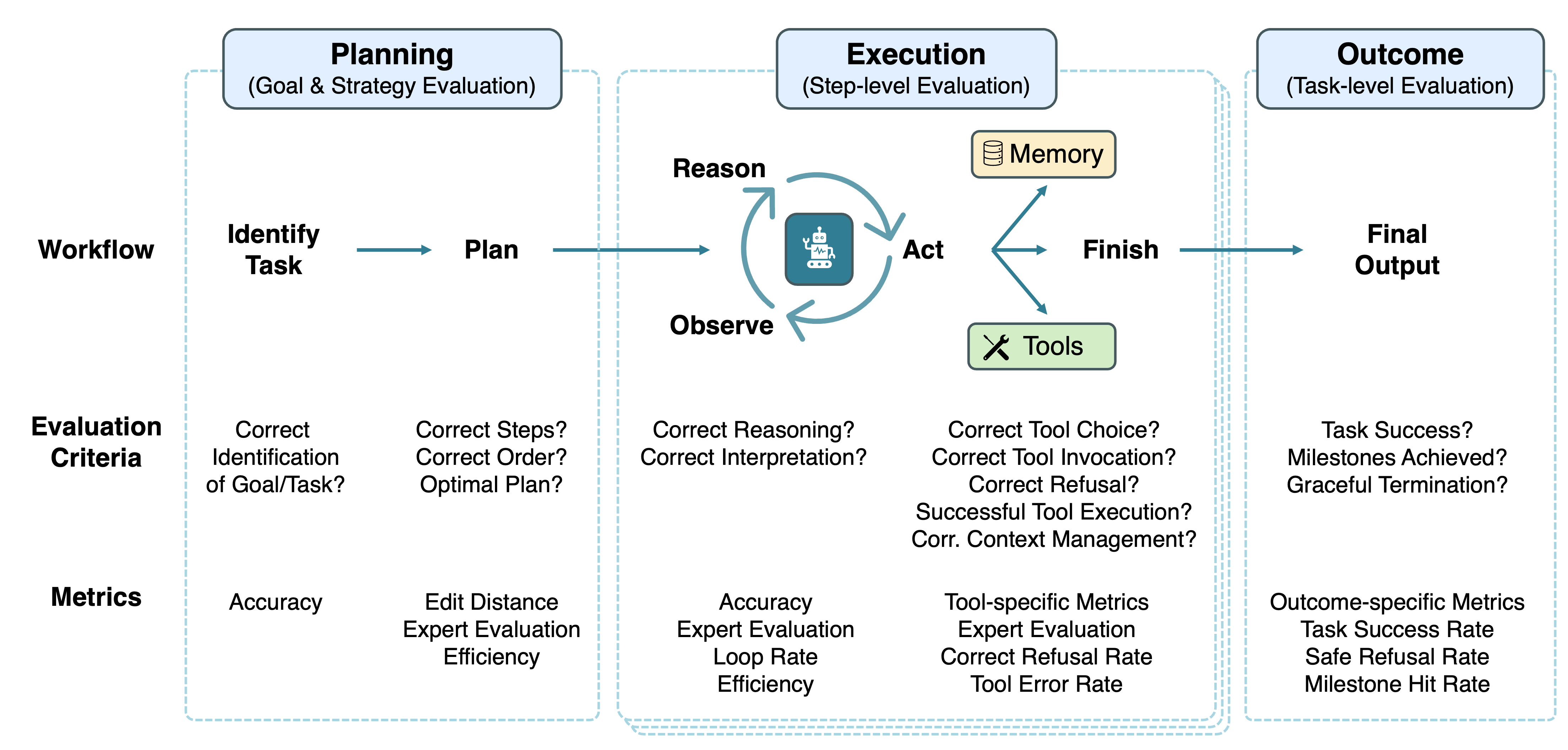}
    \caption{\textbf{High-level Evaluation Framework for Agentic Workflows}. This framework decomposes agent behavior into four tiers: Planning, Execution, Outcome and System-level evaluation (not shown). Planning assesses task identification and strategy formulation; Execution evaluates reasoning and decision-making, tool use, and memory management at each step of iterative cycles; and Outcome measures overall task success and termination quality. Note this figure omits system-level performance evaluation (e.g., costs, long-term effects) for clarity.}
    \label{fig:agent_eval}
\end{figure*}

Radiology agents navigate complex environments where they interpret open-ended queries, plan actions, adapt when results are unexpected, while still delivering useful outputs. Metrics like AUROC that suffice for narrow AI systems (e.g., CAD for pneumothorax detection) cannot capture this process. Comprehensive evaluation must therefore consider planning, execution, outcomes, and system-level performance (\figref{fig:agent_eval}).

\subsection{Planning}
\label{sec:eval_planning}

Planning begins with identifying the task and outlining a logical sequence of steps. Evaluation assesses whether the agent understood the request, proposed an appropriate plan, and avoided unnecessary steps. Because downstream reliability depends on plan clarity, ambiguity in goals or task structure can lead to execution errors; well-specified plans reduce uncertainty and allow the agent to act confidently~\cite{amodei2016concrete,yao2023react}. Replanning is equally important; if required tools are unavailable, effective agents adapt without losing sight of the goal or terminate when the task is \textit{a priori} impossible. \markup{Planning evaluation also considers responses to redundant, suboptimal, or incomplete tool palettes~\cite{zheng_can_2024}.}

Plans can be compared to expert or strong LLM ("oracle") reference plans\markup{~using distance metrics (similarity of proposed vs. reference plans) or stepwise matching accuracy, especially for order-sensitive tasks (e.g., segmentation before volume calculation)}. Expert review can also judge whether chosen actions were necessary, helpful, and reliable, particularly when multiple tools overlap~\cite{ferber2025development}. \markup{However, for more complex or dynamically changing tasks, it can be difficult to determine an "optimal" plan or even what to compare against in advance.}

\subsection{Execution}
\label{sec:eval_execution}

Execution \markup{is the process of carrying out the planned steps. For autonomous agents, this involves iterative cycles of reasoning, acting, and observing, while in structured workflows it often means following predefined sequences with constrained decision points. Step-level} evaluation assesses reasoning accuracy, action quality, and appropriate use of tools and memory.

Tool evaluation checks whether the correct tool was used at the right time with appropriate inputs, whether the tool functioned as intended, and whether outputs were correctly interpreted. Robust execution also requires handling missing or ambiguous data and issuing safe refusals when a step is impossible or unsafe. For example, if a multiphase study is missing a crucial series, the agent should clarify whether the series was omitted from PACS or adjust the plan, rather than looping or hallucinating results. \markup{An agent may still succeed despite imperfect intermediate steps, or fail after a sequence of sound ones.}

\markup{Memory use is evaluated by whether contextually relevant information (e.g., allergies or prior diagnoses) was retrieved and maintained across updates~\cite{singh2025agenticrag,wu2025longeval,li2024optimus}.}

Metrics include correctness of intermediate outputs, tool calls, and memory interactions; milestone hit rate for partial progress~\cite{zheng_can_2024}; refusal rates; loop frequency; and efficiency in actions, time, and resources. Expert assessment can additionally judge whether the reasoning and action sequence was logical~\cite{ferber2025development}.

\subsection{Outcome}
\label{sec:eval_outcome}

Outcome evaluation asks if the task was completed correctly and safely\markup{, regardless of how efficient or circuitous the process was}. Beyond simple task success rate, many task-specific metrics apply (e.g., Dice score for segmentation performance).~\cite{maier2024metrics,sai2022survey,xiaolan2025evaluating}. For close-ended tasks (with definite solutions) this may be exact accuracy against a reference standard\markup{~(although such a standard can be hard to establish in radiology)}; for open-ended tasks (with more than one valid solution) like report generation it may be expert (or LLM) rating of output quality (for instance based on similarity compared to a reference output), freedom from hallucinated findings, appropriateness of recommendations and criticality of errors. \markup{Outcomes can also be assessed for reliability across repeated runs~\cite{yao2024tau}, for instance via \textit{pass@k} (i.e., the probability that at least one of $k$ attempts succeeds) or \textit{pass}$^k$ (i.e., the probability that all $k$ attempts succeed, thereby capturing consistent rather than occasional success for critical tasks\cite{schmid2025pass}), calibration of uncertainty, and the agent’s ability to terminate gracefully (i.e., without causing additional problems) when no satisfactory solution exists (e.g., essential tools are unavailable).}

\subsection{System-level Performance}
\label{sec:eval_system}

System-level evaluation examines how agents perform beyond successful and efficient completion of workflows. It focuses on downstream effects such as radiologist efficiency~\cite{hong2025radiology,huang2025jama,chenCheXagentFoundationModel2024}, workflow integration, long-term reliability and safety, and patient outcomes. It also considers how human–agent collaboration affects cognitive load and diagnostic performance, whether biases are exposed or amplified, and risks of clinician deskilling. Broader dimensions include robustness to distribution shifts, agentic improvement over time, and resource demands like compute and energy\cite{doo2024environmental}. 

These evaluations depend on real-world deployments or realistic simulations that capture interactions among agents, IT systems, clinicians, staff, and patients, but such benchmarks currently remain underdeveloped.

\subsection{Conclusions for comprehensive evaluation}
\label{sec:eval_comprehensive}

Comprehensive evaluation integrates planning, execution, outcome, and system-level performance. No current medical agent benchmark spans all tiers, though several address individual levels (\tabref{tab:benchmarks}). RadABench is a notable example of a comprehensive benchmark for agentic radiology AI, with fine-grained evaluation of planning quality and adherence, and success and robustness of tool use, including in MAS\cite{zheng_can_2024}.

Despite their widespread use, static benchmarks such as USMLE-style multiple-choice questions offer a limited view of agent performance~\cite{raji2025benchingbenchmarks}. These tasks often assess isolated knowledge recall, rather than dynamic clinical reasoning\cite{gu2025illusion}. They are often publicly available and may have been seen during training, which compromises their validity. As such, they are best interpreted as partial indicators within a broader framework.

Given the adaptability of LLM-based agents, evaluation should move beyond static benchmarks to interactive, multi-task, multi-modal simulations that mimic real clinical complexities including noise, missing data, and ambiguities. This extends to stress testing through repeated runs, perturbations, and toolset limitations to reveal weaknesses in robustness and recovery.

Inspiration may come from objective structured clinical examinations (OSCEs) in medical education, which use standardized patients, scenarios, and multimedially enriched cases in timed stations, completed in an interactive way and scored by experts with structured checklists and global rating scales. The Sequential Diagnosis Benchmark~\cite{nori2025sequential} treats diagnosis as an iterative interaction and measures intermediate decision quality, final accuracy, and resource use. A similar benchmark could be developed for radiology agents.

Guidelines can help structure evaluations. CLAIM~\cite{tejani_checklist_2024} provides a radiology-specific reporting checklist for transparent AI studies; DEAL~\cite{tripathi2025deal} outlines best practices for developing, evaluating, and assessing LLMs in medicine (mentioning agents); TRIPOD-LLM~\cite{gallifant2025tripodllm} extends transparent reporting to LLM studies with emphasis on data provenance, human oversight, and reproducibility; and DECIDE-AI~\cite{vasey2022decideai} focuses on early-stage clinical evaluation (pre-deployment) with attention to human factors, safety, and real-world performance. While these frameworks support evaluation of LLMs, agent-specific guidelines are still lacking~\cite{mehandru2024evaluating}.

\definecolor{lightgray}{RGB}{240,240,240}

\newcolumntype{Y}{>{\centering\arraybackslash}m{0.6cm}}

\begin{table*}[t]
\centering
\scriptsize
\renewcommand{\arraystretch}{1.15}
\setlength{\tabcolsep}{4pt}
\begin{tabularx}{\textwidth}{p{0.6cm}p{2.2cm}c c c X}
\toprule
\textbf{Year} & \textbf{Benchmark}                         & \textbf{Multi-Agent} & \textbf{Planning} & \textbf{Execution} & \textbf{Summary} \\
\midrule
2024 & CRAFT-MD~\cite{bench_johri2024craftmd}                 & \cmark & \textemdash & \cmark & Simulated conversational clinical reasoning (Diagnosis, (V)QA, history taking) \\ 
2024 & AgentClinic~\cite{schmidgallAgentClinicMultimodalAgent2024}        & \cmark & \textemdash & \cmark & Conversational, multimodal diagnosis (Diagnosis, (V)QA) \\ 
2024 & MIMIC-CDM~\cite{bench_bani2025language}                   & \textemdash & \cmark & \cmark & CDM (diagnosis, treatment recommendation) for abdominal pathologies \\ 
2024 & MedChain~\cite{bench_liu2024medchain}                   & \cmark & \textemdash & \cmark & CDM (referral, history+exam, diagnosis, treatment) with 12k EHR cases \\
2024 & RadABench~\cite{zheng_can_2024}                       & \cmark & \cmark & \cmark & Radiology tasks with fine-grained plan and tool evaluation \\
2024 & SDBench~\cite{nori2025sequential}                   & (\cmark)& \cmark & (\cmark)& Sequential CDM (diagnosis by asking questions \& ordering tests) with cost evaluation  \\ 
2025 & MedAgentBench~\cite{bench_jiang2025medagentbench}         & \textemdash & \textemdash & (\cmark)& EHR query and ordering tasks in FHIR environment \\ 
2025 & MedAgentBoard~\cite{bench_zhu2025medagentboard}          & \cmark & \textemdash & (\cmark)& Single or MAS for medical QA, summarization, prediction and workflow automation \\ 
2025 & MedAgentsBench~\cite{bench_tang2025medagentsbench}        & (\cmark)& \textemdash & \textemdash & Multi-step clinical reasoning for diagnosis, with cost evaluation \\ 

\bottomrule
\end{tabularx}
\caption{
\textbf{Medical agent benchmarks.} Benchmarks were selected for their medical and radiological scope and offering sufficient reproducibility. All benchmarks evaluate for task success; none of the included benchmarks evaluates for system-level performance beyond cost evaluation. \textit{Multi-Agent}: explicit inclusion of multi-agent systems. \textit{Planning}: explicit evaluation of planning quality beyond task success (e.g., comparison with a reference plan, human review). \textit{Execution}: explicit evaluation of the execution quality beyond task success (e.g., tool-call quality, step-wise failure analysis). Symbols: \cmark full, (\cmark) partial/implicit, \textemdash not present. Abbreviations: CDM: clinical decision-making. EHR: electronic health record. FHIR: Fast Healthcare Interoperability Resources. (V)QA: (visual) question answering.
}
\label{tab:benchmarks}
\end{table*}

\section{Challenges}
\label{sec:challenges}

LLM-driven workflows and agents expand what radiology AI can do but introduce challenges beyond single-turn LLMs (Table~\ref{tab:challenges}).

\paragraph{LLM core limits.}
Agents inherit fundamental limitations of their LLM backbone: stochasticity, confabulations, bias, and poor confidence calibration as discussed in other works\cite{akincidantonoliLargeLanguageModels2023,bluethgen2025best}. Context enrichment (through RAG, memory and tool use) alleviates but does not eliminate these risks entirely. \markup{In multimodal agents, cross-modal reasoning (e.g., between a CXR and text) can be brittle, and intransparent, especially if decision traces are not inspectable.}

\paragraph{Cascading errors and context volatility.}
More than static models, agents are vulnerable to cascading errors and context degradation, where propagated inaccuracies compound across many turns.\markup{~Consider a radiologist reports a "12 mm part-solid ground-glass nodule" (recommendation CT in 3-6 months), agent $A$ changes this to "12 mm ground-glass nodule" (CT in 6-12 months), and agent $B$ reports "12 mm pulmonary opacity" (ambiguous finding, no clear recommendation). Robust step-wise validation (e.g., "Are all relevant nodule characteristics reported?") can mitigate these risks.} 

\paragraph{Multi-agent coordination.}
Multi-agent coordination presents additional complexity through resource contention and communication failures. In radiology, these issues could manifest as contradictory outcomes, duplicated work, or omission of critical steps. \markup{Bedi et al. found that optimizing individual components in multi-agent systems paradoxically reduced overall performance due to impaired information flow and inter-agent compatibility\cite{bedi2025optimization}, underscoring the importance of system-level design and validation for agentic AI applications in radiology.}

\paragraph{Integration, governance and human-AI interaction.}
Integration of agentic AI into radiology requires more than interoperability with existing systems: It raises security, regulatory, and governance concerns that existing frameworks are not fully equipped to handle. Agentic systems introduce new cybersecurity vulnerabilities like prompt injection attacks\cite{akinci2025cybersecurity,ostermann2025cybersecurity}.

Their autonomy and adaptability also reduce predictability and human oversight. Agents can act without timely intervention, increasing the risk of error propagation and unexpected behavior. As they adapt to new cases, their behavior may drift, complicating validation and undermining prior regulatory approvals. This makes it essential to define clear boundaries for agent behavior and identify when human review is required.

Most medical device regulations were designed for static, narrow-scope software and are poorly suited to autonomous, adaptive agents~\cite{freyer2025overcoming}. Proposals to address this gap include staged approvals, predefined update protocols, regulatory sandboxes, and outcome-based evaluation.

The dynamics of human–AI interaction and the effects on high-level outcomes like radiologist efficiency or patient health remain underexplored. Early studies show potential for agent-led tumor boards~\cite{ferber2025development}, diagnostic collaboration~\cite{tu2024towards}, and improved reporting efficiency with LLM-assisted workflows~\cite{hong2025radiology,chenCheXagentFoundationModel2024}, but also raise concerns about clinician deskilling~\cite{Budzyn2025deskilling}. Integrating AI agents into practice requires clarifying shared responsibilities and decision-making~\cite{rajpurkar2025beyond}, and addressing human AI-interaction biases resulting in overreliance on, or distrust in AI outputs\cite{kocakBiasArtificialIntelligence2024}.


\definecolor{lightgray}{RGB}{240,240,240}

\begin{table*}[htbp]
\centering
\scriptsize
\renewcommand{\arraystretch}{1.2}
\setlength{\tabcolsep}{4pt}
\begin{tabularx}{\textwidth}{XXXX}
\toprule
\textbf{Failure Mode} & \textbf{Risk / Impact} & \textbf{Clinical Example} & \textbf{Mitigation} \\
\midrule

\rowcolor{lightgray} \multicolumn{4}{>{\columncolor{lightgray}}l}{\textbf{Level: Data Grounding — What does the agent get to see?}} \\

Outdated, incomplete, or ambiguous input & Misguided decisions & Missing biopsy report leads to broader differential & Input validation; RAG optimization; human review \\

\rowcolor{lightgray} \multicolumn{4}{>{\columncolor{lightgray}}l}{\textbf{Level: LLM — Where can the model fail?}} \\

Biased or insufficient internal knowledge & Misinterpretation; flawed, potentially unsafe output & Agent reports "hyperdense" lesion on MRI instead of "hyperintense" & RAG with trusted sources; fine-tuning; bias audits \\

Confabulation & Plausible but false output & Agent cites a non-existent guideline & Response validation; RAG \\

Crossmodal reasoning error in multimodal LLMs & Mismatch between visual information and text output  & Agent invents a nodule or flips left/right & Modality-specific tuning; hand-off to validated vision models; expert review \\

Role or goal misunderstanding & Workflow errors; misaligned actions & Agent conducts broad literature review when asked for a specific guideline reference  & Goal validation; clear role constraints \\

Tool misuse or failure & Incorrect output or missed findings & Agent misreads failed CT tool output as "no nodules" & Task-specific validation; stress testing \\

Unrecognized uncertainty & Missed escalation to human review & Liver mass flagged as malignant without alerting radiologist & Confidence thresholds; escalation channels; fallback rules \\

\rowcolor{lightgray} \multicolumn{4}{>{\columncolor{lightgray}}l}{\textbf{Level: Execution — Where can the process fail?}} \\

Context degradation & Loss or drift of information; incorrect reasoning & Prior cancer history lost mid-task & Prioritized context; sliding-window memory; retrieval refresh \\

Cascading errors & Compounded failures & Confabulated lesion → wrong guideline → wrong recommendation & Fixed validation checkpoints; rollback; self-reflection \\

Opaque reasoning trace & Reduced possibility to audit or debug & Postulating metastatic disease without providing evidence & Structured logs; provenance tracking; explainability tools \\

Multi-agent miscoordination & Redundancy or conflict & Two agents write conflicting findings into report & Arbiter agent; task quotas; A2A protocols \\

Emergent misbehavior & Unintended/unforeseen actions & Agent cancels scheduled exams to save time & Execution sandbox; autonomy limits; active monitoring; human-in-the-loop confirmation for critical steps \\

\rowcolor{lightgray} \multicolumn{4}{>{\columncolor{lightgray}}l}{\textbf{Level: Environment \& Humans — What happens in the real world?}} \\

Poor IT integration & Broken workflows & AI report fails to transfer to PACS due to format mismatch; radiologist re-dictates manually & Validated interfaces (e.g. FHIR) to existing components  \\

(Novel) Security vulnerabilities & Attack surface increases & Compromised RAG source injects misleading info that gets interpreted as prompt & Vendor vetting; layered security; prompt defense \\

Cross-department silos & Incomplete information & Histopathology information not accessible to agent preparing MDT case vignette & Unified system access; interdepartmental integration \\

Unclear human–AI roles & Overreliance (automation bias), mistrust (algorithmic aversion bias); Deskilling & Radiologist misses fracture after AI says "normal" & Confidence calibration; training; explainability; safety roles \\

Limited external validation & Poor generalization & System underperforms at new hospital & Diverse benchmarks; prospective trials \\

Unclear accountability & Legal risk & Malpractice claim in AI-involved workflow & Clear roles; audit logs; liability protocols \\

Regulatory drift & Certification gaps & Pipeline updated without regulatory notice & Gap analysis; QMS integration \\

High environmental cost & Sustainability concerns & Weekly retraining on full PACS archive & Green compute targets; workload monitoring \\

\bottomrule
\end{tabularx}
\caption{Failure modes and mitigation strategies across layers of LLM-based radiology agents. LLM: Large Language Model; 
RAG: Retrieval-Augmented Generation;  
A2A: Agent-to-Agent (protocol); 
PACS: Picture Archiving and Communication System; 
FHIR: Fast Healthcare Interoperability Resources; 
MDT: Multidisciplinary Team; 
QMS: Quality Management System.}
\label{tab:challenges}
\end{table*}

\section{Conclusion}
\label{sec:conclusion}

LLM-driven agentic systems offer radiology a path from single-step assistance toward adaptive, multi-step automation. By offloading repetitive, non-critical tasks that contribute to cognitive load, the promise is to help radiologists refocus on high-value work, ideally at the peak of their competence. Realizing this vision requires more than technical implementation: future work must develop clinically relevant, holistic benchmarks that evaluate system-level effects; ensure robust integration with systems and human stakeholders; and rethink human-AI interaction to balance oversight, trust, and collaboration. With careful design and governance, agentic AI can evolve from experimental prototypes into valuable assistants, helping radiology adapt to rising demands while maintaining and potentially elevating quality.

\end{multicols}

\newpage
\markup{
\subsection*{Acknowledgments}
\textbf{C.B.} received research support from the Promedica Foundation, Switzerland.
\textbf{D.V.V.} is an employee of HOPPR, which develops AI technology for radiology applications.
\textbf{D.T.} received honoraria for lectures by Bayer, GE, Roche, Astra Zenica, and Philips and holds shares in StratifAI GmbH, Germany and in Synagen GmbH, Germany. D.T. is supported by the German Ministry of Research, Technology and Space (TRANSFORM LIVER - 031L0312C, SWAG - 01KD2215B, DECIPHER-M - 01KD2420B), Deutsche Forschungsgemeinschaft (DFG) (515639690), and the European Union (Horizon Europe, ODELIA - GA 101057091, ERC Starting Grant SAGMA – GA 101222556).
\textbf{J.N.K.} declares consulting services for AstraZeneca, MultiplexDx, Panakeia, Mindpeak, Owkin, DoMore Diagnostics, and Bioptimus. Furthermore, he holds shares in StratifAI, Synagen, and Spira Labs, has received an institutional research grant from GSK, and has received honoraria from AstraZeneca, Bayer, Daiichi Sankyo, Eisai, Janssen, Merck, MSD, BMS, Roche, Pfizer, and Fresenius.
\textbf{A.C.} receives research support from NIH grants R01 HL167974, R01HL169345, R01 AR077604, R01 EB002524, R01 AR079431, P41 EB027060, P50HD118632; Advanced Research Projects Agency for Health (ARPA-H) contracts AY2AX000045 and 1AYSAX0000024-01; and the Medical Imaging and Data Resource Center (MIDRC), which is funded by the National Institute of Biomedical Imaging and Bioengineering (NIBIB) under contract 75N92020C00021 and through ARPA-H. Unrelated to this work, A.C. receives research support from GE Healthcare, Philips, Microsoft, Amazon, Google, NVIDIA, Stability; has provided consulting services to Patient Square Capital, Chondrometrics GmbH, and Elucid Bioimaging; is co-founder of Cognita; has equity interest in Cognita, Subtle Medical, LVIS Corp, Brain Key.
\textbf{C.P.L:} This work was supported in part by the Medical Imaging and Data Resource Center (MIDRC), which is funded by the National Institute of Biomedical Imaging and Bioengineering (NIBIB) under contract 75N92020C00021 and through the Advanced Research Projects Agency for Health (ARPA-H). Research support from National Institutes of Health (NIH), National Institute for Biomedical Imaging and Bioengineering (NIBIB) and National Heart, Lung and Blood Institute (NHLBI), National Cancer Institute (NCI), Agency for Health Research and Quality (AHRQ); Advanced Research Projects Agency for Health (ARPA-H), and the Gordon and Betty Moore Foundation; consulting fees from Sixth Street and Gilmartin Capital; Honoraria for Lectures from Singapore Ministry of Health, Philips Medical, Canon Medical, McKinsey \& Company; Support for Travel from Singapore Ministry of Health, Philips Medica, McKinsey \& Company, C.P.L reports Patents and Patents Pending: Generalizable Machine Learning Medical Protocol Recommendation. Collaboration with GE Healthcare, India; Provision Patent Application No. 202141024621, filed November 25, 2021; Medical Autoencoders and Image Compression. Disclosure submitted with collaborators from Stanford, August 7, 2024; A System Based on a Large Language Model That Generates Sets of Patient-Centered Materials to Explain Radiology Report Information, Disclosure submitted July 24, 2025. Fiduciary duties: Board of directors, Bunkerhill Health; Board of directors Sirona Medical; Board of directors and President, Radiological Society of North America. Equity Shareholder, Bunkerhill Health, Sirona Medical; Option holder of Whiterabbit.ai, Galileo CDS, ADRA.ai, Cognita, TurboRadiology; Grants and Gifts: Grants and gifts to my institution, department, and/or research center from AWS, BunkerHill Health, Carestream, CARPL.ai, Clairity, GE HealthCare, Google Cloud, IBM, Kheiron, Lambda, Lunit, Microsoft, Nightingale Open Science, Philips, Siemens Healthineers, Stability.ai, Subtle Medical, VinBrain, Visiana, Whiterabbit.ai. Royalties: Data set license fee: annalise.ai
\textbf{F.N.} received research support from the Digitalization Initiative of the Zurich Higher Education Institutions (DIZH)- Rapid Action Call - under the TRUST-RAD project and Project Call - under the RADICAL project, and from the EU Horizon Europe programme: DataGEMS (GA.101188416).
All other authors declare no relevant conflicts of interest or acknowledgments. 

\subsection*{Author Contributions}
C.B. and F.N. conceptualized the project and created the first draft of the manuscript. C.B., M.P., D.T., T.F. and C.P.L. provided radiological perspectives. D.V.V., J.N.K., F.N., M.M., A.C., C.P.L. and M.K. provided technical advice. D.V.V., M.P., D.T., J.N.K., M.M., A.C., T.F., F.N. and C.P.L. critically revised the draft.
}

\newpage
\bibliographystyle{unsrt}
\bibliography{main}

\begin{thebibliography}{100}

\bibitem{kwee2025workload}
Thomas~C. Kwee and Robert~M. Kwee.
\newblock Workload of diagnostic radiologists in the foreseeable future based on recent (2024) scientific advances: Updated growth expectations.
\newblock {\em European Journal of Radiology}, 187:112103, June 2025.

\bibitem{yahyavi2025ajr}
Noushin Yahyavi-Firouz-Abadi.
\newblock {Preserving the Academic Mission Amid Radiologist Shortages and Financial Pressures}.
\newblock {\em American Journal of Roentgenology}, January 2025.

\bibitem{rozenshtein2024ajr}
Anna Rozenshtein, Laura~K. Findeiss, Monica~J. Wood, George Shih, and Jay~R. Parikh.
\newblock {The U.S. Radiologist Workforce: AJR Expert Panel Narrative Review}.
\newblock {\em American Journal of Roentgenology}, December 2024.

\bibitem{han2024randomised}
Ryan Han, Juli{\'a}n~N Acosta, Zahra Shakeri, John P~A Ioannidis, Eric~J Topol, and Pranav Rajpurkar.
\newblock Randomised controlled trials evaluating artificial intelligence in clinical practice: a scoping review.
\newblock {\em The Lancet Digital Health}, 6(5):e367--e373, May 2024.

\bibitem{bhayana2024chatbots}
Rajesh Bhayana.
\newblock Chatbots and {Large} {Language} {Models} in {Radiology}: {A} {Practical} {Primer} for {Clinical} and {Research} {Applications}.
\newblock {\em Radiology}, 310(1):e232756, January 2024.

\bibitem{van_veen_adapted_2024}
Dave Van~Veen, Cara Van~Uden, Louis Blankemeier, Jean-Benoit Delbrouck, Asad Aali, Christian Bluethgen, Anuj Pareek, Malgorzata Polacin, Eduardo~Pontes Reis, Anna Seehofnerová, Nidhi Rohatgi, Poonam Hosamani, William Collins, Neera Ahuja, Curtis~P. Langlotz, Jason Hom, Sergios Gatidis, John Pauly, and Akshay~S. Chaudhari.
\newblock Adapted large language models can outperform medical experts in clinical text summarization.
\newblock {\em Nature Medicine}, pages 1--9, February 2024.

\bibitem{Tu2025towardsconversational}
Tao Tu, Mike Schaekermann, Anil Palepu, Khaled Saab, Jan Freyberg, Ryutaro Tanno, Amy Wang, Brenna Li, Mohamed Amin, Yong Cheng, Elahe Vedadi, Nenad Tomasev, Shekoofeh Azizi, Karan Singhal, Le~Hou, Albert Webson, Kavita Kulkarni, S.~Sara Mahdavi, Christopher Semturs, Juraj Gottweis, Joelle Barral, Katherine Chou, Greg~S. Corrado, Yossi Matias, Alan Karthikesalingam, and Vivek Natarajan.
\newblock {Towards conversational diagnostic artificial intelligence}.
\newblock {\em Nature}, 642(8067):442--450, June 2025.

\bibitem{McDuff2025towardsddx}
Daniel McDuff, Mike Schaekermann, Tao Tu, Anil Palepu, Amy Wang, Jake Garrison, Karan Singhal, Yash Sharma, Shekoofeh Azizi, Kavita Kulkarni, Le~Hou, Yong Cheng, Yun Liu, S.~Sara Mahdavi, Sushant Prakash, Anupam Pathak, Christopher Semturs, Shwetak Patel, Dale~R. Webster, Ewa Dominowska, Juraj Gottweis, Joelle Barral, Katherine Chou, Greg~S. Corrado, Yossi Matias, Jake Sunshine, Alan Karthikesalingam, and Vivek Natarajan.
\newblock {Towards Accurate Differential Diagnosis with Large Language Models}.
\newblock {\em Nature}, 642(8067):451--457, April 2025.

\bibitem{weng_llm_nodate}
Lilian Weng.
\newblock {LLM Powered Autonomous Agents}.
\newblock \url{https://lilianweng.github.io/posts/2023-06-23-agent/}, 2023.
\newblock Accessed 11 April 2025.

\bibitem{russel2020aimodernapproach}
Stuart Russell and Peter Norvig.
\newblock {\em {Artificial Intelligence: A Modern Approach (4th Edition)}}.
\newblock Pearson, 2020.

\bibitem{effagents2024anthropic}
Erik Schluntz and Barry Zhang.
\newblock {Building Effective Agents}.
\newblock \url{https://www.anthropic.com/engineering/building-effective-agents}, 2025.
\newblock Accessed: 2025-03-10.

\bibitem{xi2025rise}
Zhiheng Xi, Wenxiang Chen, Xin Guo, Wei He, Yiwen Ding, Boyang Hong, Ming Zhang, Junzhe Wang, Senjie Jin, Enyu Zhou, Rui Zheng, Xiaoran Fan, Xiao Wang, Limao Xiong, Yuhao Zhou, Weiran Wang, Changhao Jiang, Yicheng Zou, Xiangyang Liu, Zhangyue Yin, Shihan Dou, Rongxiang Weng, Wensen Cheng, Qi~Zhang, Wenjuan Qin, Yongyan Zheng, Xipeng Qiu, Xuanjing Huang, and Tao Gui.
\newblock The rise and potential of large language model based agents: A survey.
\newblock {\em Science China Information Sciences}, 68(2):121101, 2025.

\bibitem{sutton2018reinforcement}
Richard~S. Sutton and Andrew~G. Barto.
\newblock {\em {Reinforcement Learning: An Introduction}}.
\newblock Adaptive Computation and Machine Learning series. MIT Press, Cambridge, MA, 2nd edition, 2018.

\bibitem{silver2021reward}
David Silver, Satinder Singh, Doina Precup, and Richard~S. Sutton.
\newblock Reward is enough.
\newblock {\em Artificial Intelligence}, 299:103535, 2021.

\bibitem{yu2021reinforcement}
Chao Yu, Jiming Liu, Shamim Nemati, and Guosheng Yin.
\newblock {Reinforcement Learning in Healthcare: A Survey}.
\newblock {\em ACM Computing Surveys (CSUR)}, 55(1):1--36, 2021.

\bibitem{bluethgen2025best}
Christian Bluethgen, Dave Van~Veen, Cyril Zakka, Katherine~E Link, Aaron~Hunter Fanous, Roxana Daneshjou, Thomas Frauenfelder, Curtis~P Langlotz, Sergios Gatidis, and Akshay Chaudhari.
\newblock {Best Practices for Large Language Models in Radiology}.
\newblock {\em Radiology}, 315(1):e240528, April 2025.

\bibitem{paschali2025foundation}
Magdalini Paschali, Zhihong Chen, Louis Blankemeier, Maya Varma, Alaa Youssef, Christian Bluethgen, Curtis Langlotz, Sergios Gatidis, and Akshay Chaudhari.
\newblock {Foundation Models in Radiology: What, How, Why, and Why Not}.
\newblock {\em Radiology}, 314(2):e240597, February 2025.

\bibitem{ouyangTrainingLanguageModels2022}
Long Ouyang, Jeff Wu, Xu~Jiang, Diogo Almeida, Carroll~L. Wainwright, Pamela Mishkin, Chong Zhang, Sandhini Agarwal, Katarina Slama, Alex Ray, John Schulman, Jacob Hilton, Fraser Kelton, Luke Miller, Maddie Simens, Amanda Askell, Peter Welinder, Paul Christiano, Jan Leike, and Ryan Lowe.
\newblock {Training Language Models to Follow Instructions with Human Feedback}.
\newblock In {\em Advances in Neural Information Processing Systems}, volume~35, pages 27730--27744. Curran Associates, Inc., 2022.

\bibitem{brown2020language}
Tom Brown, Benjamin Mann, Nick Ryder, Melanie Subbiah, Jared~D. Kaplan, Prafulla Dhariwal, Arvind Neelakantan, Pranav Shyam, Girish Sastry, Amanda Askell, Sandhini Agarwal, Ariel Herbert-Voss, Gretchen Krueger, Tom Henighan, Rewon Child, Aditya Ramesh, Daniel~M. Ziegler, Jeffrey Wu, Clemens Winter, Christopher Hesse, Mark Chen, Eric Sigler, Mateusz Litwin, Scott Gray, Benjamin Chess, Jack Clark, Christopher Berner, Sam McCandlish, Alec Radford, Ilya Sutskever, and Dario Amodei.
\newblock {Language Models are Few-Shot Learners}.
\newblock In H.~Larochelle, M.~Ranzato, R.~Hadsell, M.~F. Balcan, and H.~Lin, editors, {\em Advances in Neural Information Processing Systems}, volume~33, pages 1877--1901. Curran Associates, Inc., 2020.

\bibitem{weiChainofThoughtPromptingElicits2023}
Jason Wei, Xuezhi Wang, Dale Schuurmans, Maarten Bosma, Brian Ichter, Fei Xia, Ed~H. Chi, Quoc~V. Le, and Denny Zhou.
\newblock {Chain-of-Thought Prompting Elicits Reasoning in Large Language Models}.
\newblock In {\em Advances in Neural Information Processing Systems 35 (NeurIPS 2022)}, pages 24824--24837. Curran Associates, Inc., 2022.

\bibitem{hazra2025largelanguagemodelsreason}
Rishi Hazra, Gabriele Venturato, Pedro Zuidberg Dos~Martires, and Luc De~Raedt.
\newblock {Have Large Language Models Learned to Reason? A Characterization via 3-SAT Phase Transition}.
\newblock {\em arXiv preprint arXiv:2504.03930}, 2025.

\bibitem{webb2024improvingplanninglargelanguage}
Taylor Webb, Shanka~Subhra Mondal, and Ida Momennejad.
\newblock {A Brain-inspired Agentic Architecture to Improve Planning with {LLMs}}.
\newblock {\em Nature Communications}, 16(1):8633, 2025.

\bibitem{kambhampati2024llmscantplanhelp}
Subbarao Kambhampati, Karthik Valmeekam, Lin Guan, Mudit Verma, Kaya Stechly, Siddhant Bhambri, Lucas~Paul Saldyt, and Anil~B Murthy.
\newblock {Position: {LLM}s Can't Plan, But Can Help Planning in {LLM}-Modulo Frameworks}.
\newblock In Ruslan Salakhutdinov, Zico Kolter, Katherine Heller, Adrian Weller, Nuria Oliver, Jonathan Scarlett, and Felix Berkenkamp, editors, {\em Proceedings of the 41st International Conference on Machine Learning}, volume 235 of {\em Proceedings of Machine Learning Research}, pages 22895--22907. PMLR, July 2024.

\bibitem{schickToolformerLanguageModels2023}
Timo Schick, Jane Dwivedi-Yu, Roberto Dess{\`{\i}}, Roberta Raileanu, Maria Lomeli, Eric Hambro, Luke Zettlemoyer, Nicola Cancedda, and Thomas Scialom.
\newblock {Toolformer: Language Models Can Teach Themselves to Use Tools}.
\newblock In {\em Advances in Neural Information Processing Systems}, volume~36 of {\em NeurIPS}, 2023.

\bibitem{wang2024exploring}
Yiqi Wang, Wentao Chen, Xiaotian Han, Xudong Lin, Haiteng Zhao, Yongfei Liu, Bohan Zhai, Jianbo Yuan, Quanzeng You, and Hongxia Yang.
\newblock {Exploring the Reasoning Abilities of Multimodal Large Language Models (MLLMs): A Comprehensive Survey on Emerging Trends in Multimodal Reasoning}.
\newblock {\em CoRR}, abs/2401.06805, January 2024.

\bibitem{nooralahzadeh2024explainablemultimodaldataexploration}
Farhad Nooralahzadeh, Yi~Zhang, Jonathan F{\"u}rst, and Kurt Stockinger.
\newblock {Explainable Multi-Modal Data Exploration in Natural Language via LLM Agent}.
\newblock {\em arXiv preprint arXiv:2412.18428}, 2024.

\bibitem{pwc_ai_agent_survey_2025}
{PwC}.
\newblock Pwc’s {AI} agent survey.
\newblock \url{https://www.pwc.com/us/en/tech-effect/ai-analytics/ai-agent-survey.html}, 2025.
\newblock Accessed: 2025-10-07.

\bibitem{harnad1990grounding}
Stevan Harnad.
\newblock The symbol grounding problem.
\newblock {\em Physica D: Nonlinear Phenomena}, 42(1):335--346, 1990.

\bibitem{bajaj2023grounding}
Goonmeet Bajaj, Srinivasan Parthasarathy, Valerie~L. Shalin, and Amit Sheth.
\newblock {Grounding From an AI and Cognitive Science Lens}.
\newblock {\em IEEE Intelligent Systems}, 39(2):66--71, 2024.

\bibitem{liu2023grounding}
Bing Liu.
\newblock {Grounding for Artificial Intelligence}.
\newblock {\em arXiv preprint arXiv:2312.09532}, 2023.

\bibitem{xiao2024visualgrounding}
Linhui Xiao, Xiaoshan Yang, Xiangyuan Lan, Yaowei Wang, and Changsheng Xu.
\newblock {Towards Visual Grounding: A Survey}.
\newblock {\em arXiv preprint arXiv:2412.20206}, 2024.

\bibitem{huyen2025ai}
Chip Huyen.
\newblock {\em {AI Engineering: Building Applications with Foundation Models}}.
\newblock O'Reilly Media, 2025.

\bibitem{yang_survey_2025}
Yingxuan Yang, Huacan Chai, Yuanyi Song, Siyuan Qi, Muning Wen, Ning Li, Junwei Liao, Haoyi Hu, Jianghao Lin, Gaowei Chang, Weiwen Liu, Ying Wen, Yong Yu, and Weinan Zhang.
\newblock {A Survey of AI Agent Protocols}.
\newblock {\em arXiv preprint arXiv:2504.16736}, 2025.

\bibitem{sumers2023cognitive}
Theodore~R. Sumers, Shunyu Yao, Karthik Narasimhan, and Thomas~L. Griffiths.
\newblock {Cognitive Architectures for Language Agents}.
\newblock {\em Transactions on Machine Learning Research}, 2024.

\bibitem{langchain2025}
Harrison Chase and the LangChain~Contributors.
\newblock {LangChain}: A framework for developing applications with language models.
\newblock \url{https://github.com/langchain-ai/langchain}, 2025.

\bibitem{khattab2024dspy}
Omar Khattab, Arnav Singhvi, Paridhi Maheshwari, Zhiyuan Zhang, Keshav Santhanam, Sri~A. Vardhamanan, Saiful Haq, Ashutosh Sharma, Thomas~T. Joshi, Hanna Moazam, Heather Miller, Matei Zaharia, and Christopher Potts.
\newblock {DSPy: Compiling Declarative Language Model Calls into Self-Improving Pipelines}.
\newblock In {\em Proceedings of the Twelfth International Conference on Learning Representations}, 2024.

\bibitem{smolagents2025}
Hugging Face.
\newblock smolagents: a barebones library for agents that think in code.
\newblock \url{https://github.com/huggingface/smolagents}, 2025.
\newblock Accessed 2025‑07‑06.

\bibitem{woelflein2025toolmaker}
Georg Wölflein, Dyke Ferber, Daniel Truhn, Ognjen Arandjelović, and Jakob~Nikolas Kather.
\newblock {LLM Agents Making Agent Tools}.
\newblock In {\em Proceedings of the 63rd Annual Meeting of the Association for Computational Linguistics (Volume 1: Long Papers)}, pages 26092--26130, Vienna, Austria, 2025. Association for Computational Linguistics.

\bibitem{amodei2016concrete}
Dario Amodei, Chris Olah, Jacob Steinhardt, Paul Christiano, John Schulman, and Dan Man{\'e}.
\newblock {Concrete Problems in AI Safety}.
\newblock {\em arXiv preprint arXiv:1606.06565}, 2016.

\bibitem{krakovna2020specification}
Victoria Krakovna, Jonathan Uesato, Vladimir Mikulik, Matthew Rahtz, Tom Everitt, Ramana Kumar, Zac Kenton, Jan Leike, and Shane Legg.
\newblock {Specification Gaming: The Flip Side of AI Ingenuity}.
\newblock DeepMind Blog, \url{https://deepmind.google/discover/blog/specification-gaming-the-flip-side-of-ai-ingenuity/}, April 2020.
\newblock Accessed 1 September 2025.

\bibitem{xu2025visual}
Yi~Xu, Chengzu Li, Han Zhou, Xingchen Wan, Caiqi Zhang, Anna Korhonen, and Ivan Vuli{\'c}.
\newblock {Visual Planning: Let's Think Only with Images}.
\newblock may 2025.
\newblock arXiv:2505.11409, v2 last revised 2025-09-29.

\bibitem{hao_training_2024}
Shibo Hao, Sainbayar Sukhbaatar, DiJia Su, Xian Li, Zhiting Hu, Jason Weston, and Yuandong Tian.
\newblock Training {Large} {Language} {Models} to {Reason} in a {Continuous} {Latent} {Space}.
\newblock {\em arXiv}, December 2024.
\newblock arXiv:2412.06769v2 [cs.CL], revised 2024-12-11.

\bibitem{geiping2025scalingtesttimecomputelatent}
Jonas Geiping, Sean McLeish, Neel Jain, John Kirchenbauer, Siddharth Singh, Brian~R. Bartoldson, Bhavya Kailkhura, Abhinav Bhatele, and Tom Goldstein.
\newblock {Scaling up Test-Time Compute with Latent Reasoning: A Recurrent Depth Approach}, 2 2025.
\newblock arXiv preprint.

\bibitem{wang2023selfconsistency}
Xuezhi Wang, Jason Wei, Dale Schuurmans, Quoc~V. Le, Ed~H. Chi, Sharan Narang, Aakanksha Chowdhery, and Denny Zhou.
\newblock {Self-Consistency Improves Chain of Thought Reasoning in Language Models}.
\newblock In {\em The Eleventh International Conference on Learning Representations (ICLR 2023)}, 2023.

\bibitem{yao2023tree}
Shunyu Yao, Dian Yu, Jeffrey Zhao, Izhak Shafran, Thomas~L. Griffiths, Yuan Cao, and Karthik Narasimhan.
\newblock {Tree of Thoughts: Deliberate Problem Solving with Large Language Models}.
\newblock In {\em Advances in Neural Information Processing Systems}, volume~36, pages 11809--11822, 2023.

\bibitem{koh2024tree}
Jing~Yu Koh, Stephen McAleer, Daniel Fried, and Ruslan Salakhutdinov.
\newblock {Tree Search for Language Model Agents}.
\newblock {\em Transactions on Machine Learning Research}, 2025.

\bibitem{openai_o3_2025}
{OpenAI}.
\newblock Introducing {OpenAI} {o3} and {o4-mini}.
\newblock \url{https://openai.com/index/introducing-o3-and-o4-mini}, April 2025.
\newblock OpenAI Blog, accessed 3 July 2025.

\bibitem{Guo2025DeepSeek}
Daya Guo, Dejian Yang, Haowei Zhang, Junxiao Song, Ruoyu Zhang, Runxin Xu, Qihao Zhu, Shirong Ma, Peiyi Wang, Xiao Bi, et~al.
\newblock {DeepSeek\textendash R1} incentivizes reasoning in {LLMs} through reinforcement learning.
\newblock {\em Nature}, 645(8081):633--638, 2025.

\bibitem{yao2023react}
Shunyu Yao, Jeffrey Zhao, Dian Yu, Nan Du, Izhak Shafran, Karthik Narasimhan, and Yuan Cao.
\newblock {ReAct}: {S}ynergizing {R}easoning and {A}cting in {L}anguage {M}odels.
\newblock In {\em The Eleventh International Conference on Learning Representations}, 2023.

\bibitem{xu2023rewoo}
Binfeng Xu, Zhiyuan Peng, Bowen Lei, Subhabrata Mukherjee, Yuchen Liu, and Dongkuan Xu.
\newblock {ReWOO}: {Decoupling} {Reasoning} from {Observations} for {Efficient} {Augmented} {Language} {Models}.
\newblock {\em arXiv preprint arXiv:2305.18323}, 2023.

\bibitem{shinn2023reflexion}
Noah Shinn, Federico Cassano, Ashwin Gopinath, Karthik~R. Narasimhan, and Shunyu Yao.
\newblock {Reflexion: Language Agents with Verbal Reinforcement Learning}.
\newblock In {\em Thirty-seventh Conference on Neural Information Processing Systems}, 2023.

\bibitem{schmid2024context}
Philipp Schmid.
\newblock {The New Skill in AI is Not Prompting, It's Context Engineering}.
\newblock \url{https://www.philschmid.de/context-engineering}, June 2025.
\newblock Accessed 3 July 2025.

\bibitem{mei2025contextengineering}
Lingrui Mei, Jiayu Yao, Yuyao Ge, Yiwei Wang, Baolong Bi, Yujun Cai, Jiazhi Liu, Mingyu Li, Zhong-Zhi Li, Duzhen Zhang, Chenlin Zhou, Jiayi Mao, Tianze Xia, Jiafeng Guo, and Shenghua Liu.
\newblock {A Survey of Context Engineering for Large Language Models}.
\newblock {\em arXiv preprint arXiv:2507.13334}, 2025.

\bibitem{li2024promptcompression}
Zongqian Li, Yinhong Liu, Yixuan Su, and Nigel Collier.
\newblock {Prompt Compression for Large Language Models: A Survey}.
\newblock In {\em Proceedings of the 2025 Conference of the Nations of the Americas Chapter of the Association for Computational Linguistics: Human Language Technologies (Volume 1: Long Papers)}, pages 7182--7195, Albuquerque, New Mexico, 2025. Association for Computational Linguistics.

\bibitem{zakkaAlmanacRetrievalAugmentedLanguage2024}
Cyril Zakka, Rohan Shad, Akash Chaurasia, Alex~R. Dalal, Jennifer~L. Kim, Michael Moor, Robyn Fong, Curran Phillips, Kevin Alexander, Euan Ashley, Jack Boyd, Kathleen Boyd, Karen Hirsch, Curt Langlotz, Rita Lee, Joanna Melia, Joanna Nelson, Karim Sallam, Stacey Tullis, Melissa~Ann Vogelsong, John~Patrick Cunningham, and William Hiesinger.
\newblock Almanac — {Retrieval}-{Augmented} {Language} {Models} for {Clinical} {Medicine}.
\newblock {\em NEJM AI}, 1(2):AIoa2300068, 2024.

\bibitem{zheng2025miriad}
Qinyue Zheng, Salman Abdullah, Sam Rawal, Cyril Zakka, Sophie Ostmeier, Maximilian Purk, Eduardo Reis, Eric~J. Topol, Jure Leskovec, and Michael Moor.
\newblock {MIRIAD: Augmenting LLMs with Millions of Medical Query-Response Pairs}.
\newblock {\em arXiv preprint arXiv:2506.06091}, 2025.

\bibitem{arasteh2024radiorag}
Soroosh~Tayebi Arasteh, Mahshad Lotfinia, Keno Bressem, Robert Siepmann, Lisa Adams, Dyke Ferber, Christiane Kuhl, Jakob~Nikolas Kather, Sven Nebelung, and Daniel Truhn.
\newblock {RadioRAG: Online Retrieval-Augmented Generation for Radiology Question Answering}.
\newblock {\em Radiology: Artificial Intelligence}, 7(4):e240476, 2025.

\bibitem{truhn2023large}
Daniel Truhn, Jorge~S. Reis-Filho, and Jakob~Nikolas Kather.
\newblock {Large language models should be used as scientific reasoning engines, not knowledge databases}.
\newblock {\em Nature Medicine}, 29(12):2983--2984, 2023.

\bibitem{hong2025contextrot}
Kelly Hong, Anton Troynikov, and Jeff Huber.
\newblock {Context Rot: How Increasing Input Tokens Impacts LLM Performance}.
\newblock \url{https://research.trychroma.com/context-rot}, July 2025.

\bibitem{liAgentHospitalSimulacrum2024a}
Junkai Li, Yunghwei Lai, Weitao Li, Jingyi Ren, Meng Zhang, Xinhui Kang, Siyu Wang, Peng Li, Ya-Qin Zhang, Weizhi Ma, and Yang Liu.
\newblock {Agent Hospital: A Simulacrum of Hospital with Evolvable Medical Agents}.
\newblock {\em arXiv preprint arXiv:2405.02957}, 2024.

\bibitem{schmidgallAgentClinicMultimodalAgent2024}
Samuel Schmidgall, Rojin Ziaei, Carl Harris, Eduardo Reis, Jeffrey Jopling, and Michael Moor.
\newblock {AgentClinic}: A multimodal agent benchmark to evaluate {{AI}} in simulated clinical environments.
\newblock {\em arXiv preprint arXiv:2405.07960}, 2025.

\bibitem{gao2025selfevolve}
Huan-ang Gao, Jiayi Geng, Wenyue Hua, Mengkang Hu, Xinzhe Juan, Hongzhang Liu, Shilong Liu, Jiahao Qiu, Xuan Qi, Yiran Wu, Hongru Wang, Han Xiao, Yuhang Zhou, Shaokun Zhang, Jiayi Zhang, Jinyu Xiang, Yixiong Fang, Qiwen Zhao, Dongrui Liu, Qihan Ren, Cheng Qian, Zhenghailong Wang, Minda Hu, Huazheng Wang, Qingyun Wu, Heng Ji, and Mengdi Wang.
\newblock {A Survey of Self-Evolving Agents: On Path to Artificial Super Intelligence}.
\newblock {\em arXiv preprint arXiv:2507.21046}, 2025.

\bibitem{chen2024spin}
Zixiang Chen, Yihe Deng, Huizhuo Yuan, Kaixuan Ji, and Quanquan Gu.
\newblock {Self-Play Fine-Tuning Converts Weak Language Models to Strong Language Models}.
\newblock In {\em Proceedings of the 41st International Conference on Machine Learning}, volume 235 of {\em Proceedings of Machine Learning Research}, pages 6621--6642. PMLR, Jul 2024.

\bibitem{zhou2025self}
Yifei Zhou, Sergey Levine, Jason Weston, Xian Li, and Sainbayar Sukhbaatar.
\newblock {Self-Challenging Language Model Agents}.
\newblock {\em arXiv preprint arXiv:2506.01716}, jun 2025.

\bibitem{kimSeeingUnseenAdvancing2024}
Woojin Kim.
\newblock Seeing the {Unseen}: {Advancing} {Generative} {AI} {Research} in {Radiology}.
\newblock {\em Radiology}, 311(2):e240935, May 2024.

\bibitem{wu2022ai}
Tongshuang Wu, Michael Terry, and Carrie~J. Cai.
\newblock {AI} {Chains}: {Transparent} and {Controllable} {Human}-{AI} {Interaction} by {Chaining} {Large} {Language} {Model} {Prompts}.
\newblock In {\em Proceedings of the 2022 CHI Conference on Human Factors in Computing Systems (CHI '22)}, New York, NY, USA, 2022. Association for Computing Machinery.

\bibitem{wu2024autogen}
Qingyun Wu, Gagan Bansal, Jieyu Zhang, Yiran Wu, Beibin Li, Erkang Zhu, Li~Jiang, Xiaoyun Zhang, Shaokun Zhang, Jiale Liu, Ahmed~Hassan Awadallah, Ryen~W White, Doug Burger, and Chi Wang.
\newblock {AutoGen}: Enabling {Next-Gen} {LLM} applications via {Multi-Agent} conversations.
\newblock In {\em Proceedings of the First Conference on Language Modeling (COLM)}, Philadelphia, PA, USA, October 2024.

\bibitem{zhang2025agentorchestra}
Wentao Zhang, Liang Zeng, Yuzhen Xiao, Yongcong Li, Ce~Cui, Yilei Zhao, Rui Hu, Yang Liu, Yahui Zhou, and Bo~An.
\newblock {AgentOrchestra: Orchestrating Hierarchical Multi-Agent Intelligence with the Tool-Environment-Agent(TEA) Protocol}.
\newblock {\em arXiv preprint arXiv:2506.12508}, 2025.

\bibitem{chen2024autoagents}
Guangyao Chen, Siwei Dong, Yu~Shu, Ge~Zhang, Jaward Sesay, B{\"o}rje Karlsson, Jie Fu, and Yemin Shi.
\newblock {AutoAgents: A Framework for Automatic Agent Generation}.
\newblock In Kate Larson, editor, {\em {Proceedings of the Thirty-Third International Joint Conference on Artificial Intelligence, {{IJCAI-24}}}}, pages 22--30, 2024.

\bibitem{jimenez2025swarm}
Cristian Jimenez-Romero, Alper Yegenoglu, and Christian Blum.
\newblock {Multi-agent systems powered by large language models: applications in swarm intelligence}.
\newblock {\em Frontiers in Artificial Intelligence}, 8:1593017, 2025.

\bibitem{huang2025biomni}
Kexin Huang, Serena Zhang, Hanchen Wang, Yuanhao Qu, Yingzhou Lu, Yusuf Roohani, Ryan Li, Lin Qiu, Gavin Li, Junze Zhang, Di~Yin, Shruti Marwaha, Jennefer~N. Carter, Xin Zhou, Matthew Wheeler, Jonathan~A. Bernstein, Mengdi Wang, Peng He, Jingtian Zhou, Michael Snyder, Le~Cong, Aviv Regev, and Jure Leskovec.
\newblock {Biomni: A General-Purpose Biomedical AI Agent}.
\newblock {\em bioRxiv}, 2025.

\bibitem{zeng2024radcouncil}
Fang Zeng, Zhiliang Lyu, Quanzheng Li, and Xiang Li.
\newblock {Enhancing LLMs for Impression Generation in Radiology Reports through a Multi-Agent System}.
\newblock {\em arXiv preprint arXiv:2412.06828}, 2024.

\bibitem{wasserthal2023totalsegmentator}
Jakob Wasserthal, Hanns-Christian Breit, Manfred~T Meyer, Maurice Pradella, Daniel Hinck, Alexander~W Sauter, Tobias Heye, Daniel~T Boll, Joshy Cyriac, Shan Yang, et~al.
\newblock Totalsegmentator: robust segmentation of 104 anatomic structures in ct images.
\newblock {\em Radiology: Artificial Intelligence}, 5(5):e230024, 2023.

\bibitem{chenCheXagentFoundationModel2024}
Zhihong Chen, Maya Varma, Justin Xu, Magdalini Paschali, Dave Van~Veen, Andrew Johnston, Alaa Youssef, Louis Blankemeier, Christian Bluethgen, Stephan Altmayer, Jeya Maria~Jose Valanarasu, Mohamed Siddig~Eltayeb Muneer, Eduardo~Pontes Reis, Joseph~Paul Cohen, Cameron Olsen, Tanishq~Mathew Abraham, Emily~B. Tsai, Christopher~F. Beaulieu, Jenia Jitsev, Sergios Gatidis, Jean-Benoit Delbrouck, Akshay~S. Chaudhari, and Curtis~P. Langlotz.
\newblock {A Vision-Language Foundation Model to Enhance Efficiency of Chest X-ray Interpretation}.
\newblock {\em arXiv preprint arXiv:2401.12208}, 2024.

\bibitem{deperrois2025radvlm}
Nicolas Deperrois, Hidetoshi Matsuo, Samuel Ruip{\'e}rez-Campillo, Moritz Vandenhirtz, Sonia Laguna, Alain Ryser, Koji Fujimoto, Mizuho Nishio, Thomas~M. Sutter, Julia~E. Vogt, Jonas Kluckert, Thomas Frauenfelder, Christian Bl{\"u}thgen, Farhad Nooralahzadeh, and Michael Krauthammer.
\newblock {RadVLM: A Multitask Conversational Vision-Language Model for Radiology}.
\newblock {\em arXiv preprint arXiv:2502.03333}, 2025.

\bibitem{hamamci_developing_2025}
Ibrahim~Ethem Hamamci, Sezgin Er, Chenyu Wang, Furkan Almas, Ayse~Gulnihan Simsek, Sevval~Nil Esirgun, Irem Dogan, Omer~Faruk Durugol, Benjamin Hou, Suprosanna Shit, Weicheng Dai, Murong Xu, Hadrien Reynaud, Muhammed~Furkan Dasdelen, Bastian Wittmann, Tamaz Amiranashvili, Enis Simsar, Mehmet Simsar, Emine~Bensu Erdemir, Abdullah Alanbay, Anjany Sekuboyina, Berkan Lafci, Ahmet Kaplan, Zhiyong Lu, Malgorzata Polacin, Bernhard Kainz, Christian Bluethgen, Kayhan Batmanghelich, Mehmet~Kemal Ozdemir, and Bjoern Menze.
\newblock {Developing Generalist Foundation Models from a Multimodal Dataset for 3D Computed Tomography}.
\newblock {\em arXiv preprint arXiv:2403.17834}, 2025.

\bibitem{blankemeier2024merlin}
Louis Blankemeier, Joseph~Paul Cohen, Ashwin Kumar, Dave Van~Veen, Syed Jamal~Safdar Gardezi, Magdalini Paschali, Zhihong Chen, Jean-Benoit Delbrouck, Eduardo Reis, Cesar Truyts, Christian Bluethgen, Malte Engmann Kjeldskov~Jensen, Sophie Ostmeier, Maya Varma, Jeya Maria~Jose Valanarasu, Zhongnan Fang, Zepeng Huo, Zaid Nabulsi, Diego Ardila, Wei-Hung Weng, Edson Amaro, Neera Ahuja, Jason Fries, Nigam~H Shah, Andrew Johnston, Robert~D Boutin, Andrew Wentland, Curtis~P Langlotz, Jason Hom, Sergios Gatidis, and Akshay~S Chaudhari.
\newblock {Merlin: A Vision Language Foundation Model for 3D Computed Tomography}.
\newblock {\em Research Square}, pages rs.3.rs--4546309, 2024.
\newblock Preprint.

\bibitem{ferber2025development}
Dyke Ferber, Omar~SM El~Nahhas, Georg W{\"o}lflein, Isabella~C Wiest, Jan Clusmann, Marie-Elisabeth Le{\ss}mann, Sebastian Foersch, Jacqueline Lammert, Maximilian Tschochohei, Dirk J{\"a}ger, et~al.
\newblock Development and validation of an autonomous artificial intelligence agent for clinical decision-making in oncology.
\newblock {\em Nature cancer}, pages 1--13, 2025.

\bibitem{zheng_can_2024}
Qiaoyu Zheng, Chaoyi Wu, Pengcheng Qiu, Lisong Dai, Ya~Zhang, Yanfeng Wang, and Weidi Xie.
\newblock {Can Modern LLMs Act as Agent Cores in Radiology Environments?}
\newblock {\em arXiv preprint arXiv:2412.09529}, 2024.

\bibitem{chen2025radfabric}
Wenting Chen, Yi~Dong, Zhaojun Ding, Yucheng Shi, Yifan Zhou, Fang Zeng, Yijun Luo, Tianyu Lin, Yihang Su, Yichen Wu, Kai Zhang, Zhen Xiang, Tianming Liu, Ninghao Liu, Lichao Sun, Yixuan Yuan, and Xiang Li.
\newblock {RadFabric: Agentic AI System with Reasoning Capability for Radiology}.
\newblock {\em arXiv preprint arXiv:2506.14142}, 2025.

\bibitem{an_bi-rads_2019}
Julie~Y. An, Kyle M.~L. Unsdorfer, and Jeffrey~C. Weinreb.
\newblock {BI}-{RADS}, {C}-{RADS}, {CAD}-{RADS}, {LI}-{RADS}, {Lung}-{RADS}, {NI}-{RADS}, {O}-{RADS}, {PI}-{RADS}, {TI}-{RADS}: {Reporting} and {Data} {Systems}.
\newblock {\em RadioGraphics}, 39(5):1435--1436, September 2019.
\newblock Publisher: Radiological Society of North America.

\bibitem{uicc2025tnm9}
James~D. Brierley, Meredith Giuliani, Brian O’Sullivan, Brian Rous, and Elizabeth Van~Eycken, editors.
\newblock {\em TNM Classification of Malignant Tumours}.
\newblock John Wiley \& Sons, Oxford, 9th edition, July 2025.

\bibitem{eisenhauer2009recist1_1}
Elizabeth~A. Eisenhauer, Paul Therasse, Jan Bogaerts, et~al.
\newblock New response evaluation criteria in solid tumours: Revised recist guideline (version 1.1).
\newblock {\em European Journal of Cancer}, 45(2):228--247, 2009.

\bibitem{kather2024llminterface}
Jakob~Nikolas Kather, Dyke Ferber, Isabella~C Wiest, Stephen Gilbert, and Daniel Truhn.
\newblock Large language models could make natural language again the universal interface of healthcare.
\newblock {\em Nature Medicine}, 30(10):2708--2710, 2024.

\bibitem{filice_integrating_2019}
Ross~W. Filice and Charles~E. Kahn.
\newblock Integrating an {Ontology} of {Radiology} {Differential} {Diagnosis} with {ICD}-10-{CM}, {RadLex}, and {SNOMED} {CT}.
\newblock {\em Journal of Digital Imaging}, 32(2):206--210, April 2019.

\bibitem{chepelev_ontologies_2023}
Leonid~L. Chepelev, David Kwan, Charles~E. Kahn, Ross~W. Filice, and Kenneth~C. Wang.
\newblock Ontologies in the {New} {Computational} {Age} of {Radiology}: {RadLex} for {Semantics} and {Interoperability} in {Imaging} {Workflows}.
\newblock {\em RadioGraphics}, 43(3):e220098, March 2023.

\bibitem{vreeman_loinc_2018}
Daniel~J Vreeman, Swapna Abhyankar, Kenneth~C Wang, Christopher Carr, Beverly Collins, Daniel~L Rubin, and Curtis~P Langlotz.
\newblock The {LOINC} {RSNA} radiology playbook - a unified terminology for radiology procedures.
\newblock {\em Journal of the American Medical Informatics Association}, 25(7):885--893, July 2018.

\bibitem{budovec2014informatics}
Joseph~J Budovec, Cesar~A Lam, and Charles~E Kahn~Jr.
\newblock Informatics in radiology: radiology gamuts ontology: differential diagnosis for the semantic web.
\newblock {\em Radiographics}, 34(1):254--264, 2014.

\bibitem{delbrouck2024radgraph}
Jean-Benoit Delbrouck, Pierre Chambon, Zhihong Chen, Maya Varma, Andrew Johnston, Louis Blankemeier, Dave Van~Veen, Tan Bui, Steven Truong, and Curtis Langlotz.
\newblock {RadGraph-XL: A large-scale expert-annotated dataset for entity and relation extraction from radiology reports}.
\newblock In {\em Findings of the Association for Computational Linguistics: ACL 2024}, pages 12902--12915, 2024.

\bibitem{chang2024snomed}
Eunsuk Chang and Sumi Sung.
\newblock {Use of SNOMED CT in Large Language Models: Scoping Review}.
\newblock {\em JMIR Medical Informatics}, 12(1):e62924, 2024.

\bibitem{arnold2024fhir}
Philipp Arnold, Daniel~Pinto Dos~Santos, Fabian Bamberg, and Elmar Kotter.
\newblock {FHIR--Overdue Standard for Radiology Data Warehouses}.
\newblock In {\em R{\"o}Fo-Fortschritte auf dem Gebiet der R{\"o}ntgenstrahlen und der bildgebenden Verfahren}. Georg Thieme Verlag KG, 2024.

\bibitem{ChristianHinge2025dicommcp}
Christian Hinge.
\newblock {dicom-mcp}: Model context protocol for interacting with dicom servers.
\newblock \url{https://github.com/ChristianHinge/dicom-mcp}, 2025.
\newblock Version v0.1.2 (released 2025-04-28); accessed 2025-08-01.

\bibitem{collaco2025agenticAI}
Bernardo~Gabriele Collaco, Syed~Ali Haider, Srinivasagam Prabha, Cesar~Abraham Gomez-Cabello, Ariana Genovese, Nadia~G. Wood, Sanjay~P. Bagaria, Narayanan Gopala, Cui Tao, and Antonio~Jorge Forte.
\newblock {The Role of Agentic Artificial Intelligence in Healthcare: A Systematic Review}.
\newblock {\em Research Square}, August 2025.
\newblock Preprint, Version 1.

\bibitem{strayer2023lcs}
Thomas~E Strayer, Lucy~B Spalluto, Abby Burns, Christopher~J Lindsell, Claudia~I Henschke, David~F Yankelevitz, Drew Moghanaki, Robert~S Dittus, Timothy~J Vogus, Carolyn Audet, et~al.
\newblock Using the framework for reporting adaptations and modifications-expanded (frame) to study adaptations in lung cancer screening delivery in the veterans health administration: a cohort study.
\newblock {\em Implementation science communications}, 4(1):5, 2023.

\bibitem{meng2025totalregistrator}
Xuan~Loc Pham, Gwendolyn Vuurberg, Marjan Doppen, Joey Roosen, Tip Stille, Thi~Quynh Ha, Thuy~Duong Quach, Quoc~Vu Dang, Manh~Ha Luu, Ewoud~J. Smit, Hong~Son Mai, Mattias Heinrich, Bram van Ginneken, Mathias Prokop, and Alessa Hering.
\newblock {TotalRegistrator: Towards a Lightweight Foundation Model for CT Image Registration}.
\newblock {\em arXiv preprint arXiv:2508.04450}, 2025.

\bibitem{bluethgen_visionlanguage_2024}
Christian Bluethgen, Pierre Chambon, Jean-Benoit Delbrouck, Rogier Van Der~Sluijs, Małgorzata Połacin, Juan~Manuel Zambrano~Chaves, Tanishq~Mathew Abraham, Shivanshu Purohit, Curtis~P. Langlotz, and Akshay~S. Chaudhari.
\newblock A vision–language foundation model for the generation of realistic chest {X}-ray images.
\newblock {\em Nature Biomedical Engineering}, August 2024.

\bibitem{dabass2025asrllm}
Manju Dabass, Mohammed~M. Jabeer, Anuj Chandalia, and Dwarikanath Mahapatra.
\newblock {Streamlined Speech Recognition Model for Automated Radiology Reporting Employing Combined Automatic Speech Recognition Model, Large Language Model, and Prompt Engineering}.
\newblock In Gaurav Raj, Bhuvan Unhelker, and Ankur Choudhary, editors, {\em Advances in Artificial Intelligence and Machine Learning}, pages 345--356, Singapore, 2025. Springer Nature Singapore.

\bibitem{singh2025agenticrag}
Aditi Singh, Abul Ehtesham, Saket Kumar, and Tala Talaei~Khoei.
\newblock {Agentic Retrieval-Augmented Generation: A Survey on Agentic RAG}.
\newblock {\em arXiv preprint arXiv:2501.09136}, 2025.

\bibitem{wu2025longeval}
Siwei Wu, Yizhi Li, Xingwei Qu, Rishi Ravikumar, Yucheng Li, Tyler Loakman, Shanghaoran Quan, Xiaoyong Wei, Riza Batista-Navarro, and Chenghua Lin.
\newblock {LongEval: A Comprehensive Analysis of Long-Text Generation Through a Plan-based Paradigm}.
\newblock {\em arXiv preprint arXiv:2502.19103}, 2025.

\bibitem{li2024optimus}
Zaijing Li, Yuquan Xie, Rui Shao, Gongwei Chen, Dongmei Jiang, and Liqiang Nie.
\newblock {Optimus-1: Hybrid multimodal memory empowered agents excel in long-horizon tasks}.
\newblock {\em Advances in neural information processing systems}, 37:49881--49913, 2024.

\bibitem{maier2024metrics}
Lena Maier-Hein, Annika Reinke, Patrick Godau, Minu~D Tizabi, Florian Buettner, Evangelia Christodoulou, Ben Glocker, Fabian Isensee, Jens Kleesiek, Michal Kozubek, et~al.
\newblock Metrics reloaded: recommendations for image analysis validation.
\newblock {\em Nature methods}, 21(2):195--212, 2024.

\bibitem{sai2022survey}
Ananya~B Sai, Akash~Kumar Mohankumar, and Mitesh~M Khapra.
\newblock A survey of evaluation metrics used for nlg systems.
\newblock {\em ACM Computing Surveys (CSUR)}, 55(2):1--39, 2022.

\bibitem{xiaolan2025evaluating}
Chen Xiaolan, Xiang Jiayang, Lu~Shanfu, Liu Yexin, He~Mingguang, and Shi Danli.
\newblock Evaluating large language models and agents in healthcare: key challenges in clinical applications.
\newblock {\em Intelligent Medicine}, 2025.

\bibitem{yao2024tau}
Shunyu Yao, Noah Shinn, Pedram Razavi, and Karthik Narasimhan.
\newblock $\tau$-bench: A benchmark for tool-agent-user interaction in real-world domains.
\newblock {\em arXiv preprint arXiv:2406.12045}, 2024.

\bibitem{schmid2025pass}
Philipp Schmid.
\newblock Pass@k vs pass\^k: Understanding agent reliability.
\newblock \url{https://www.philschmid.de/agents-pass-at-k-pass-power-k}, 2025.
\newblock Accessed: 2025-10-09.

\bibitem{hong2025radiology}
Eun~Kyoung Hong, Byungseok Roh, Beomhee Park, Jae-Bock Jo, Woong Bae, Jai~Soung Park, Dong-Wook Sung, et~al.
\newblock {Value of Using a Generative AI Model in Chest Radiography Reporting: A Reader Study}.
\newblock {\em Radiology}, 314(3):e241646, 2025.

\bibitem{huang2025jama}
Jonathan Huang, Matthew~T. Wittbrodt, Caitlin~N. Teague, Eric Karl, et~al.
\newblock {Efficiency and Quality of Generative AI–Assisted Radiograph Reporting}.
\newblock {\em JAMA Network Open}, 8(6):e2513921, 2025.

\bibitem{doo2024environmental}
Florence~X Doo, Jan Vosshenrich, Tessa~S Cook, Linda Moy, Eduardo~PRP Almeida, Sean~A Woolen, Judy~Wawira Gichoya, Tobias Heye, and Kate Hanneman.
\newblock {Environmental sustainability and AI in radiology: a double-edged sword}.
\newblock {\em Radiology}, 310(2):e232030, 2024.

\bibitem{raji2025benchingbenchmarks}
Inioluwa~Deborah Raji, Roxana Daneshjou, and Emily Alsentzer.
\newblock It’s time to bench the medical exam benchmark.
\newblock {\em {NEJM AI}}, 2(2):AIe2401235, 2025.

\bibitem{gu2025illusion}
Yu~Gu, Jingjing Fu, Xiaodong Liu, Jeya Maria~Jose Valanarasu, Noel Codella, Reuben Tan, Qianchu Liu, Ying Jin, Sheng Zhang, Jinyu Wang, et~al.
\newblock The illusion of readiness: Stress testing large frontier models on multimodal medical benchmarks.
\newblock {\em arXiv preprint arXiv:2509.18234}, 2025.

\bibitem{nori2025sequential}
Harsha Nori, Mayank Daswani, Christopher Kelly, Scott Lundberg, Marco~Tulio Ribeiro, Marc Wilson, Xiaoxuan Liu, Viknesh Sounderajah, Jonathan Carlson, Matthew~P Lungren, et~al.
\newblock {Sequential Diagnosis with Language Models}.
\newblock {\em arXiv preprint arXiv:2506.22405}, 2025.

\bibitem{tejani_checklist_2024}
Ali~S. Tejani, Michail~E. Klontzas, Anthony~A. Gatti, John~T. Mongan, Linda Moy, Seong~Ho Park, Charles~E. Kahn, {for the CLAIM 2024 Update Panel}, Sunhy Abbara, Saif Afat, Udunna~C. Anazodo, Anna Andreychenko, Folkert~W. Asselbergs, Aldo Badano, Bettina Baessler, Bayarbaatar Bold, Sotirios Bisdas, Torkel~B. Brismar, Giovanni~E. Cacciamani, John~A. Carrino, Julius Chapiro, Michael~F. Chiang, Tessa~S. Cook, Renato Cuocolo, John Damilakis, Roxana Daneshjou, Carlo~N. De~Cecco, Hesham Elhalawani, Guillermo Elizondo-Riojas, Andrey Fedorov, Benjamin Fine, Adam~E. Flanders, Judy Wawira~Gichoya, Maryellen~L. Giger, Safwan~S. Halabi, Sven Haller, William Hsu, Krishna Juluru, Jayashree Kalpathy-Cramer, Apostolos~H. Karantanas, Felipe~C. Kitamura, Burak Kocak, Dow-Mu Koh, Elmar Kotter, Elizabeth~A. Krupinski, Curtis~P. Langlotz, Cecilia~S. Lee, Mario Maas, Anant Madabhushi, Lena Maier-Hein, Kostas Marias, Luis Martí-Bonmatí, Jaishree Naidoo, Emanuele Neri, Robert Ochs, Nikolaos Papanikolaou, Thomas Papathomas, Katja
  Pinker-Domenig, Daniel Pinto Dos~Santos, Fred Prior, Alexandros Protonotarios, Mauricio Reyes, Veronica Rotemberg, Jeffrey~D. Rudie, Emmanuel Salinas-Miranda, Francesco Sardanelli, Mark~E. Schweitzer, Luca~Maria Sconfienza, Ronnie Sebro, Prateek Sharma, An~Tang, Antonios Tzortzakakis, Jeroen Van Der~Laak, Peter M.~A. Van~Ooijen, Vasantha~K. Venugopal, Jacob~J. Visser, Bradford~J. Wood, Carol~C. Wu, Greg Zaharchuk, and Marc Zins.
\newblock Checklist for {Artificial} {Intelligence} in {Medical} {Imaging} ({CLAIM}): 2024 {Update}.
\newblock {\em Radiology: Artificial Intelligence}, 6(4):e240300, 2024.

\bibitem{tripathi2025deal}
Satvik Tripathi, Dana Alkhulaifat, Florence~X Doo, Pranav Rajpurkar, Rafe McBeth, Dania Daye, and Tessa~S Cook.
\newblock {Development, Evaluation, and Assessment of Large Language Models (DEAL) Checklist: A Technical Report}.
\newblock {\em {NEJM AI}}, 2(6):AIp2401106, 2025.

\bibitem{gallifant2025tripodllm}
Jack Gallifant, Majid Afshar, Saleem Ameen, Yindalon Aphinyanaphongs, Shan Chen, Giovanni Cacciamani, Dina Demner-Fushman, Dmitriy Dligach, Roxana Daneshjou, Chrystinne Fernandes, et~al.
\newblock {The TRIPOD-LLM reporting guideline for studies using large language models}.
\newblock {\em Nature medicine}, 31(1):60--69, 2025.

\bibitem{vasey2022decideai}
Baptiste Vasey, Myura Nagendran, Bruce Campbell, David~A Clifton, Gary~S Collins, Spiros Denaxas, Alastair~K Denniston, Livia Faes, Bart Geerts, Mudathir Ibrahim, Xiaoxuan Liu, Basil~A Mateen, Piyush Mathur, Michael~D McCradden, Lauren Morgan, Jonathan Ordish, Charlotte Rogers, Suchi Saria, Daniel S~W Ting, Peter Watkinson, Wolf Weber, Paul Wheatstone, Peter McCulloch, and DECIDE-AI expert group.
\newblock {Reporting guideline for the early-stage clinical evaluation of decision support systems driven by artificial intelligence: DECIDE-AI}.
\newblock {\em {Nature Medicine}}, 28(5):924--933, 2022.

\bibitem{mehandru2024evaluating}
Nikita Mehandru, Brenda~Y Miao, Eduardo~Rodriguez Almaraz, Madhumita Sushil, Atul~J Butte, and Ahmed Alaa.
\newblock Evaluating large language models as agents in the clinic.
\newblock {\em NPJ digital medicine}, 7(1):84, 2024.

\bibitem{bench_johri2024craftmd}
Shreya Johri, Jaehwan Jeong, Benjamin~A. Tran, Daniel~I Schlessinger, Shannon Wongvibulsin, Zhuo~Ran Cai, Roxana Daneshjou, and Pranav Rajpurkar.
\newblock {CRAFT}-{MD}: A conversational evaluation framework for comprehensive assessment of clinical {LLM}s.
\newblock In {\em AAAI 2024 Spring Symposium on Clinical Foundation Models}, 2024.

\bibitem{bench_bani2025language}
David Bani-Harouni, Chantal Pellegrini, Ege {\"O}zsoy, Matthias Keicher, and Nassir Navab.
\newblock {Language Agents for Hypothesis-driven Clinical Decision Making with Reinforcement Learning}.
\newblock {\em arXiv preprint arXiv:2506.13474}, 2025.

\bibitem{bench_liu2024medchain}
Jie Liu, Wenxuan Wang, Zizhan Ma, Guolin Huang, Yihang Su, Kao-Jung Chang, Wenting Chen, Haoliang Li, Linlin Shen, and Michael Lyu.
\newblock {MedChain: Bridging the Gap Between LLM Agents and Clinical Practice through Interactive Sequential Benchmarking}.
\newblock {\em arXiv preprint arXiv:2412.01605}, 2024.

\bibitem{bench_jiang2025medagentbench}
Yixing Jiang, Kameron~C. Black, Gloria Geng, Danny Park, James Zou, Andrew~Y. Ng, and Jonathan~H. Chen.
\newblock {MedAgentBench: A Virtual EHR Environment to Benchmark Medical LLM Agents}.
\newblock {\em {NEJM AI}}, 2(9):AIdbp2500144, 2025.

\bibitem{bench_zhu2025medagentboard}
Yinghao Zhu, Ziyi He, Haoran Hu, Xiaochen Zheng, Xichen Zhang, Zixiang Wang, Junyi Gao, Liantao Ma, and Lequan Yu.
\newblock {MedAgentBoard: Benchmarking Multi-Agent Collaboration with Conventional Methods for Diverse Medical Tasks}.
\newblock {\em arXiv preprint arXiv:2505.12371}, 2025.

\bibitem{bench_tang2025medagentsbench}
Xiangru Tang, Daniel Shao, Jiwoong Sohn, Jiapeng Chen, Jiayi Zhang, Jinyu Xiang, Fang Wu, Yilun Zhao, Chenglin Wu, Wenqi Shi, Arman Cohan, and Mark Gerstein.
\newblock {MedAgentsBench: Benchmarking Thinking Models and Agent Frameworks for Complex Medical Reasoning}.
\newblock {\em arXiv preprint arXiv:2503.07459}, 2025.

\bibitem{akincidantonoliLargeLanguageModels2023}
Tugba Akinci~D'Antonoli, Arnaldo Stanzione, Christian Bluethgen, Federica Vernuccio, Lorenzo Ugga, Michail~E. Klontzas, Renato Cuocolo, Roberto Cannella, and Burak Koçak.
\newblock Large language models in radiology: fundamentals, applications, ethical considerations, risks, and future directions.
\newblock {\em Diagnostic and Interventional Radiology (Ankara, Turkey)}, October 2023.

\bibitem{bedi2025optimization}
Suhana Bedi, Iddah Mlauzi, Daniel Shin, Sanmi Koyejo, and Nigam~H. Shah.
\newblock {The Optimization Paradox in Clinical AI Multi-Agent Systems}.
\newblock {\em arXiv preprint arXiv:2506.06574}, 2025.

\bibitem{akinci2025cybersecurity}
Tugba Akinci~D’Antonoli, Ali~S Tejani, Bardia Khosravi, Christian Bluethgen, Felix Busch, Keno~K Bressem, Lisa~C Adams, Mana Moassefi, Shahriar Faghani, and Judy~Wawira Gichoya.
\newblock {Cybersecurity Threats and Mitigation Strategies for Large Language Models in Health Care}.
\newblock {\em {Radiology: Artificial Intelligence}}, 7(4):e240739, 2025.

\bibitem{ostermann2025cybersecurity}
Max Ostermann, Rebecca Mathias, Fatemeh Jahed, Mitchell~B Parker, Florence~D Hudson, William~C Harding, Stephen Gilbert, and Oscar Freyer.
\newblock {Cybersecurity Requirements for Medical Devices in the EU and US-A Comparison and Gap Analysis of the MDCG 2019-16 and FDA premarket cybersecurity guidance}.
\newblock {\em Computational and Structural Biotechnology Journal}, 2025.

\bibitem{freyer2025overcoming}
Oscar Freyer, Sanddhya Jayabalan, Jakob~N Kather, and Stephen Gilbert.
\newblock Overcoming regulatory barriers to the implementation of ai agents in healthcare.
\newblock {\em Nature Medicine}, pages 1--5, 2025.

\bibitem{tu2024towards}
Tao Tu, Shekoofeh Azizi, Danny Driess, Mike Schaekermann, Mohamed Amin, Pi-Chuan Chang, Andrew Carroll, Charles Lau, Ryutaro Tanno, Ira Ktena, et~al.
\newblock Towards generalist biomedical ai.
\newblock {\em {NEJM AI}}, 1(3):AIoa2300138, 2024.

\bibitem{Budzyn2025deskilling}
Krzysztof Budzyń, Marcin Romańczyk, Diana Kitala, Paweł Kołodziej, Marek Bugajski, Hans~O Adami, Johannes Blom, Marek Buszkiewicz, Natalie Halvorsen, Cesare Hassan, Tomasz Romańczyk, Øyvind Holme, Krzysztof Jarus, Shona Fielding, Melina Kunar, Maria Pellise, Nastazja Pilonis, Michał~Filip Kamiński, Mette Kalager, Michael Bretthauer, and Yuichi Mori.
\newblock Endoscopist deskilling risk after exposure to artificial intelligence in colonoscopy: a multicentre, observational study.
\newblock {\em The Lancet Gastroenterology \& Hepatology}, 2025.

\bibitem{rajpurkar2025beyond}
Pranav Rajpurkar and Eric~J Topol.
\newblock {Beyond Assistance: The Case for Role Separation in AI-Human Radiology Workflows}.
\newblock {\em Radiology}, 316(1):e250477, 2025.

\bibitem{kocakBiasArtificialIntelligence2024}
Burak Koçak, Andrea Ponsiglione, Arnaldo Stanzione, Christian Bluethgen, João Santinha, Lorenzo Ugga, Merel Huisman, Michail~E. Klontzas, Roberto Cannella, and Renato Cuocolo.
\newblock Bias in artificial intelligence for medical imaging: fundamentals, detection, avoidance, mitigation, challenges, ethics, and prospects.
\newblock {\em Diagnostic and Interventional Radiology (Ankara, Turkey)}, July 2024.

\end{thebibliography}

\newpage

\end{document}